\pdfoutput=1
\documentclass[preprints,article,accept,pdftex,moreauthors]{Definitions/mdpi}
\AtBeginDocument{\setlength{\extralength}{0pt}\setlength{\fulllength}{\textwidth}}

\firstpage{1}
\pubvolume{1}
\issuenum{1}
\articlenumber{0}
\pubyear{2026}
\copyrightyear{2026}
\datereceived{ }
\daterevised{ }
\dateaccepted{ }
\datepublished{ }
\hreflink{https://doi.org/}

\usepackage{siunitx}
\Title{From Cloud to Edge: A Fully Local Pipeline for Pothole Risk Assessment and Driver Warning Generation}

\TitleCitation{From Cloud to Edge: A Fully Local Pipeline for Pothole Risk Assessment and Driver Warning Generation}

\Author{Abdalmalek Aburaddaha * and Samir A. Rawashdeh}

\AuthorNames{Abdalmalek Aburaddaha and Samir A. Rawashdeh}

\isAPAStyle{%
       \AuthorCitation{Aburaddaha, A., \& Rawashdeh, S. A.}
         }{%
        \isChicagoStyle{%
        \AuthorCitation{Aburaddaha, Abdalmalek, and Samir A. Rawashdeh.}
        }{
        \AuthorCitation{Aburaddaha, A.; Rawashdeh, S.A.}
        }
}

\address[1]{%
Department of Electrical and Computer Engineering, University of Michigan--Dearborn, Dearborn, MI 48128, USA}

\corres{Correspondence: abdmalek@umich.edu}

\abstract{Vision--language models can turn road images into clear, geometry-aware pothole warnings, but current systems depend on commercial cloud services. Our earlier five-stage pipeline needed two cloud calls per scene, taking about 20~s and \$0.105, and could not run offline. We rebuild every stage to run on a single local GPU and evaluate it with the same data, prompts and protocol. Segmentation keeps the original Segment Anything weights but runs them in half precision with graph compilation, which makes it 2.7$\times$ faster without changing any downstream risk class; five lightweight variants could not match its masks on small road defects. Severity is graded by six ConvNeXt students distilled from 3092 cloud labels. They reach Cohen's $\kappa = 0.268$ against human consensus at 28~ms per pothole, close to the cloud model's 0.321, within the camera setup they were trained for. A local 8-billion-parameter vision--language model writes the warnings under a grammar that admits only in-lane maneuvers. Three blind, independent raters scored these warnings safer than the cloud warnings (4.18 vs.\ 3.65 of 5; $p = 0.001$; 50 scenes) and equally clear, whereas an automated LLM judge ranked them the other way. The complete pipeline runs at 1.27~s per scene, 16$\times$ faster than the cloud system, offline and at no per-scene cost.}


\begin{document}

\section{Introduction}
\label{sec:intro}

Potholes damage vehicles, injure cyclists and motorcyclists, and force sudden maneuvers that put other road users at risk~\cite{koch2011pothole,ma2022review}. Road agencies need to know where the worst defects are so they can schedule repairs, and drivers or automated vehicles need timely warnings so they can slow down or adjust their path. Both needs depend on more than detection: a system must judge \emph{how severe} each pothole is and \emph{what the driver should do about it}.

In earlier work we built a five-stage pipeline for this task~\cite{aburaddaha2026pothole}. It corrects the camera perspective, segments each pothole with the Segment Anything Model (SAM)~\cite{kirillov2023sam}, measures its shape and position, grades its severity with a commercial vision--language model (VLM), and then asks the same VLM to write a driver warning grounded in the measured geometry. Blind raters preferred these grounded warnings over image-only warnings by 3.5 to 1. The severity grades are perceived visual categories rather than depth measurements~\cite{aburaddaha2026pothole}.

Every scene, however, needed two calls to a cloud API, which took about 20~s and cost about \$0.105. For vehicles and road-survey fleets, cloud dependence has further costs: the system stops working without a network connection, hosted models can be updated or withdrawn without notice, and per-call fees grow with the size of a survey. A city-wide survey, or a warning that must reach the driver before the vehicle reaches the hazard, needs something faster and self-contained.

\begin{figure}[H]
\begin{adjustwidth}{-\extralength}{0cm}
\centering
\includegraphics[width=\linewidth]{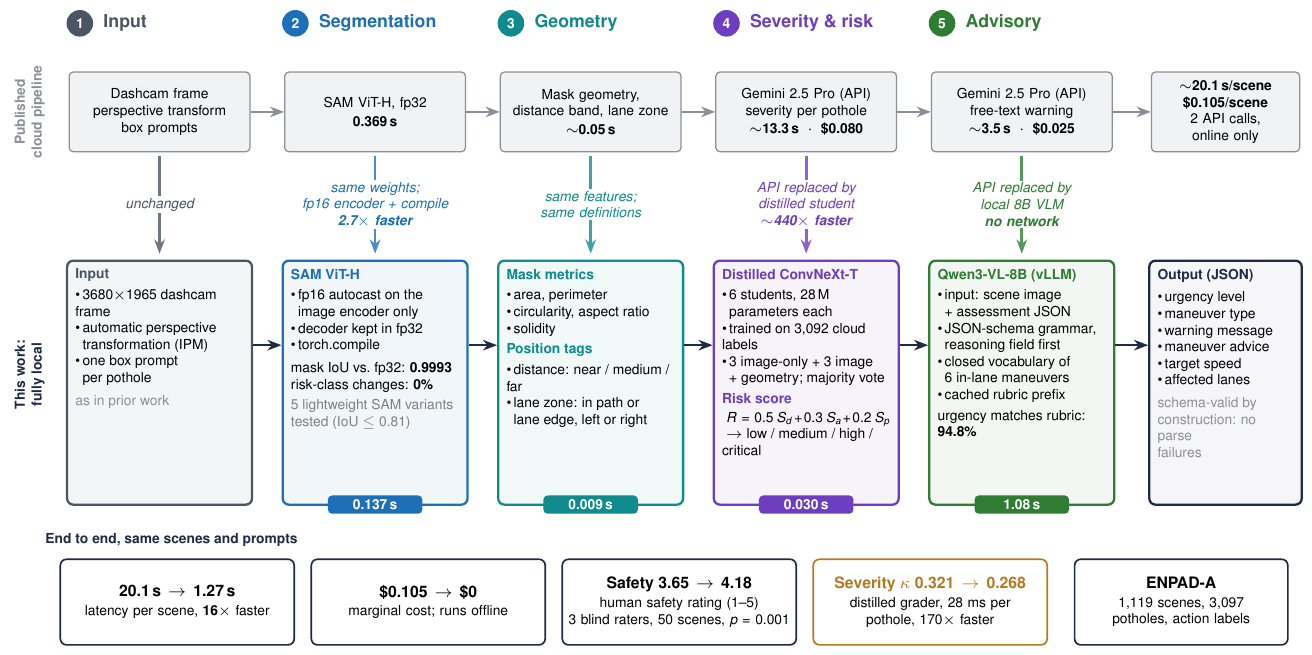}
\end{adjustwidth}
\caption{Overview of this work. Top: the published cloud pipeline~\cite{aburaddaha2026pothole}, which calls a commercial VLM twice per scene. Bottom: the fully local pipeline of this paper. Arrows in the middle state what changed at each stage. The bottom strip gives the end-to-end outcome on the same scenes and prompts. Stage times are steady-state means per scene over the 1119-scene corpus.\label{fig:blockdiagram}}
\end{figure}

This paper builds that system. We replace every cloud stage with a local one and compare the two pipelines stage by stage under two rules: the data, prompts and evaluation protocol stay the same as in the cloud system, and every replacement must meet acceptance criteria fixed before its results were seen. Figure~\ref{fig:blockdiagram} summarises the outcome.

The main contributions are as follows:
\begin{enumerate}
\item \textbf{A fully local pothole assessment and warning pipeline.} It runs at 1.27~s per scene on one workstation GPU, 16$\times$ faster than the cloud system, with no network access and no per-scene fee. In a blind study, three independent raters scored its warnings safer than the cloud system's (+0.53 points on a 5-point scale, $p = 0.001$), equally clear, and preferred them about 2 to 1.
\item \textbf{ENPAD-A}, a dataset of 1119 scenes and 3097 potholes in which every scene carries its perception record and a structured driver action, intended for vision--language--action research and available to other researchers (see the Data Availability Statement).
\end{enumerate}

The rest of the paper is organised as follows. Section~\ref{sec:related} reviews related work. Section~\ref{sec:methods} describes the pipeline, the data and the evaluation protocol. Section~\ref{sec:results} reports results stage by stage, including the human study. Section~\ref{sec:discussion} discusses what the results mean for practice and lists the limitations, and Section~\ref{sec:conclusions} concludes.

\section{Related Work}
\label{sec:related}

\subsection{Pothole Detection and Measurement}
Pothole sensing has moved through three broad families. \emph{Vibration-based} methods mount accelerometers in vehicles or smartphones and flag road segments where the signal exceeds a threshold~\cite{eriksson2008pothole,mednis2011realtime}. They are cheap and work in the dark, but they detect a pothole only after a wheel has hit it, and they cannot say how large it is. \emph{Vision-based} methods detect potholes from camera images. Early work used hand-crafted image processing, such as histogram thresholding and texture cues~\cite{koch2011pothole,nienaber2015,ouma2017fuzzy,jo2015blackbox}; later work used convolutional networks and one-stage detectors trained on large road-damage datasets~\cite{zhang2016crack,maeda2018rdd,arya2021multicountry,arya2024rdd2022,ahmed2021dilated}. \emph{Three-dimensional} methods recover depth from stereo cameras, RGB-D sensors or laser scanners, which allows the pothole's depth and volume to be measured directly~\cite{jog2012properties,moazzam2013kinect,fan2020disparity,fan2022rethinking,dhiman2020pothole,yurdakul2025yolov8}. Recent reviews summarise these families and their trade-offs~\cite{ma2022review}.

Our own earlier work showed that correcting the dashcam perspective before detection improves pothole detection for distant defects~\cite{aburaddaha2024jimaging}. The pipeline studied here starts from that corrected view. Like most of the literature above, those methods stop at detection or measurement. They do not decide how urgent a pothole is for a particular vehicle, and they do not tell the driver what to do.

\subsection{Vision--Language Models for Road and Driving Scenes}
Large VLMs pair an image encoder with a language model~\cite{radford2021clip,liu2023llava} and can answer open questions about an image. Commercial models such as Gemini~2.5~\cite{comanici2025gemini25} and open-weight models such as Qwen2.5-VL and Qwen3-VL~\cite{bai2025qwen25vl,bai2025qwen3vl} now handle detailed visual reasoning. In driving, VLMs have been used to explain vehicle actions~\cite{kim2018textual}, to answer questions about traffic scenes~\cite{sima2024drivelm,marcu2024lingoqa}, and to plan or control the vehicle end to end~\cite{xu2024drivegpt4,tian2024drivevlm}; a recent survey covers this growing area~\cite{cui2024survey}. For road infrastructure, V-RoAst tested whether VLMs can rate road safety attributes under the iRAP standard and found that they generalise to unseen classes but are weak at spatial reasoning~\cite{jongwiriyanurak2024vroast}.

Two gaps remain. First, most driving VLMs reason about whole scenes; few ground their language in measured properties of a specific hazard. Our prior pipeline addressed this by feeding geometry into the prompt~\cite{aburaddaha2026pothole}. Second, nearly all strong results rely on very large or cloud-hosted models, while vehicles and roadside units need systems that run locally with predictable latency~\cite{liu2019edge,lin2018architectural}. This paper addresses the second gap for the pothole task.

\subsection{Efficient Promptable Segmentation}
SAM produces a mask for any object given a point or box prompt, using a large ViT-H image encoder~\cite{kirillov2023sam}. Several works shrink that encoder by distillation or redesign: MobileSAM~\cite{zhang2023mobilesam}, TinySAM~\cite{shu2025tinysam}, EfficientViT-SAM~\cite{zhang2024efficientvitsam}, EdgeSAM~\cite{zhou2025edgesam}, EfficientSAM~\cite{xiong2024efficientsam} and FastSAM~\cite{zhao2023fastsam}. SAM~2 extends the design to video~\cite{ravi2024sam2}. These models report near-parity with SAM on general benchmarks such as COCO and LVIS, but studies in specialised domains show that SAM's own accuracy drops on low-contrast or unusual targets~\cite{mazurowski2023medsam,ji2024sam_not_perfect}. We test whether the lightweight variants keep their parity on small, low-contrast road defects. A second, simpler route to speed is to keep the model and change how it is executed, with reduced-precision arithmetic~\cite{micikevicius2018mixed} and graph compilation~\cite{ansel2024pytorch2}.

\subsection{Knowledge Distillation from Large Models}
Knowledge distillation trains a small student to imitate a large teacher~\cite{hinton2015distilling,gou2021kdsurvey}. A common variant uses the teacher only to label data, after which the student is trained on those pseudo-labels~\cite{xie2020noisystudent}; the same idea is now used to distil large language models into much smaller ones~\cite{hsieh2023stepbystep}. Students can match their teachers inside the training distribution, but deep networks often learn shortcuts that fail when the data shift~\cite{geirhos2020shortcut,recht2019imagenet,koh2021wilds}. Ensembles of independently trained networks give more stable predictions and a natural confidence signal~\cite{lakshminarayanan2017ensembles}. We use an ensemble of ConvNeXt students~\cite{liu2022convnext} and measure both its in-domain accuracy and its behaviour under a change of camera geometry.

\subsection{Structured Generation and Efficient Serving}
When a language model's output must be read by software, grammar-constrained decoding guarantees valid output by masking, at each step, every token that would break a given grammar or JSON schema~\cite{willard2023outlines,geng2023grammar,dong2024xgrammar}. Serving engines such as vLLM~\cite{kwon2023vllm} and SGLang~\cite{zheng2024sglang} make local inference fast through paged memory and reuse of shared prompt prefixes. Format restrictions can, however, lower reasoning accuracy~\cite{tam2024speakfreely}, and asking a model to reason before answering often helps~\cite{wei2022cot}. We design the output schema with both effects in mind.

\subsection{Evaluating Generated Text: Humans and LLM Judges}
Human ratings remain the reference for judging generated text, and their reliability is usually reported with agreement statistics such as Cohen's or Fleiss' $\kappa$, the intraclass correlation and Krippendorff's $\alpha$~\cite{cohen1960,fleiss1971,shrout1979icc,koo2016icc,hayes2007kripp}. Because human studies are slow, LLMs are increasingly used as judges~\cite{zheng2023judging,chiang2023human,chen2024mllmjudge}. Known problems include position bias~\cite{wang2024notfair}, a preference for the judge's own outputs or those of related models~\cite{panickssery2024selfpref}, and uneven agreement with humans across tasks~\cite{bavaresco2025judges}. We evaluate our warnings with both a human study and a judge.

\subsection{Summary and Positioning}
Table~\ref{tab:positioning} places this work among the main families of related systems. To our knowledge, no previous system combines pothole localisation, severity grading and a structured driver warning in a pipeline that runs fully offline and whose warnings have been rated by humans.

\begin{table}[H]
\caption{Positioning of this work. Each row is a family of related systems; a check mark means the property is typical of that family. ``Warning'' means a driver-facing message with a recommended action. ``Local'' means the system runs without a cloud service.\label{tab:positioning}}
\begin{adjustwidth}{-\extralength}{0cm}
\newcolumntype{C}{>{\centering\arraybackslash}X}
\small
\begin{tabularx}{\fulllength}{p{5.6cm}p{2.3cm}CCCCC}
\toprule
\textbf{Family} & \textbf{Input} & \textbf{Locates potholes} & \textbf{Grades severity} & \textbf{Writes warning} & \textbf{Local} & \textbf{Human-rated output} \\
\midrule
Vibration sensing~\cite{eriksson2008pothole,mednis2011realtime} & accelerometer & $\checkmark^{a}$ & -- & -- & $\checkmark$ & -- \\
Classical image processing~\cite{koch2011pothole,nienaber2015,ouma2017fuzzy} & RGB image & $\checkmark$ & -- & -- & $\checkmark$ & -- \\
Deep detectors~\cite{maeda2018rdd,arya2021multicountry,ahmed2021dilated,aburaddaha2024jimaging} & RGB image & $\checkmark$ & -- & -- & $\checkmark$ & -- \\
3D measurement~\cite{jog2012properties,moazzam2013kinect,fan2020disparity,fan2022rethinking,yurdakul2025yolov8} & stereo, RGB-D & $\checkmark$ & $\checkmark^{b}$ & -- & $\checkmark$ & -- \\
VLM road-safety rating~\cite{jongwiriyanurak2024vroast} & RGB image & -- & $\checkmark^{c}$ & -- & -- & -- \\
Driving VLMs~\cite{sima2024drivelm,xu2024drivegpt4,tian2024drivevlm,marcu2024lingoqa} & camera video & -- & -- & $\checkmark^{d}$ & partly & partly \\
Cloud grounded pipeline~\cite{aburaddaha2026pothole} & RGB image & $\checkmark$ & $\checkmark$ & $\checkmark$ & -- & $\checkmark$ \\
\textbf{This work} & RGB image & $\checkmark$ & $\checkmark$ & $\checkmark$ & $\checkmark$ & $\checkmark$ \\
\bottomrule
\end{tabularx}
\end{adjustwidth}
\noindent{\footnotesize{$^{a}$ Only after a wheel has hit the pothole. $^{b}$ Measured depth or volume. $^{c}$ Road-safety attributes, not potholes. $^{d}$ General driving explanations or answers, not hazard-specific warnings.}}
\end{table}

\section{Materials and Methods}
\label{sec:methods}

\subsection{Pipeline Overview}
\begin{figure}[p]
\begin{adjustwidth}{-\extralength}{0cm}
\centering
\includegraphics[width=\linewidth,height=0.84\textheight,keepaspectratio]{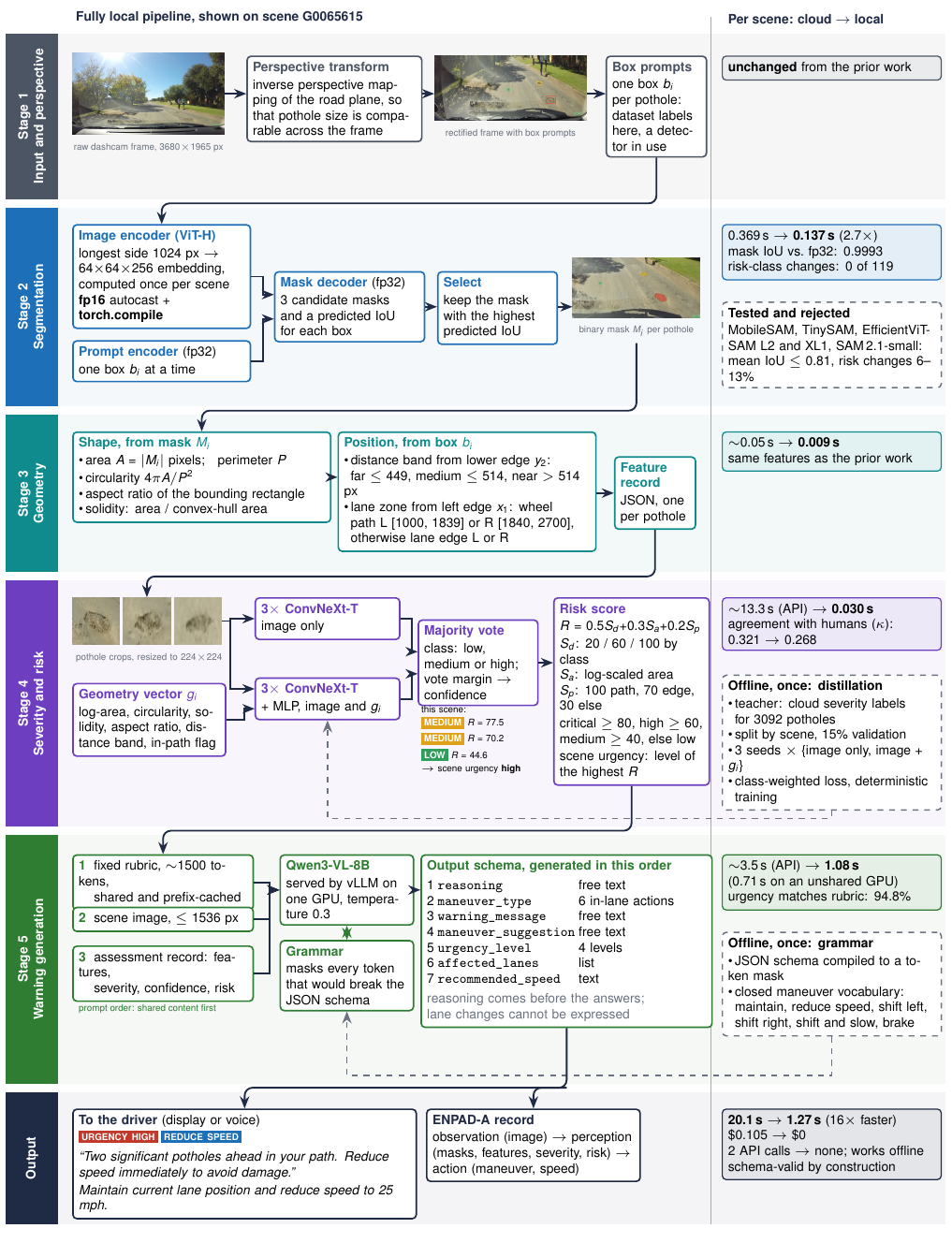}
\end{adjustwidth}
\caption{Detailed flow of the local pipeline, from the raw dashcam frame to the driver warning, shown on scene G0065615 with its actual outputs. Each band is one stage, and arrows follow the data. The right-hand column gives each stage's time per scene against the cloud pipeline, the segmentation models that were tested and rejected, and the two steps prepared offline once: distilling the severity students and compiling the output grammar.\label{fig:example}}
\end{figure}

The pipeline has five stages (Figure~\ref{fig:blockdiagram}). Stage~1 corrects the perspective of a dashcam frame. Stage~2 segments each pothole. Stage~3 measures its shape and position. Stage~4 grades its severity and combines severity, size and position into a risk score. Stage~5 writes a driver warning from the scene image and the assessment record. Figure~\ref{fig:example} shows the full flow in detail on one real scene, with its actual outputs.

This study evaluates the \emph{assessment} of potholes that have already been located. As in the cloud system, the box prompts for Stage~2 are the dataset's annotated boxes, which stand in for an upstream detector such as the one in our earlier work~\cite{aburaddaha2024jimaging}. We therefore make no claim about detection performance.

\subsection{Data}
The main corpus is the one used by the cloud system~\cite{aburaddaha2026pothole}: 1119 dashcam scenes with 3097 annotated potholes, taken from a public South African road dataset~\cite{nienaber2015} and perspective-corrected as in~\cite{aburaddaha2024jimaging}. Frames are $3680 \times 1965$ pixels. Of the 3097 potholes, 3092 carry a severity label from the cloud model (Gemini~2.5~Pro); these labels are the teacher signal for distillation in Section~\ref{sec:distill}.

Two further datasets are used for evaluation only. The first is the \emph{human consensus set} released with the prior work: 42 potholes in 12 scenes graded by human annotators, of which 38 have a usable consensus label~\cite{aburaddaha2026pothole}. All 12 scenes are excluded from every training set in this paper. The second is PothRGBD~\cite{yurdakul2025yolov8,pothrgbd_data}, an independent dataset of $640 \times 480$ RGB images with co-registered depth from an Intel RealSense sensor and pothole masks. Its close-range camera geometry differs strongly from our dashcam images, which makes it a test of how each method copes with a change of camera setup.

\subsection{Ground Rules and Acceptance Criteria}
\label{sec:rules}
To keep the comparison with the cloud system fair, prompts, scene sets and scoring code are reused without change unless a change is stated. All timings use synchronised GPU timers after a warm-up and report the steady-state mean per scene. Before replacing a stage we first reproduced the published Stage-2 masks exactly (mask IoU $= 1.000$ over 119 masks) so that we had a valid reference.

For Stage~2 we fixed two acceptance criteria before testing any candidate:
\begin{enumerate}[label=(C\arabic*),leftmargin=*]
\item at most 2\% of potholes may change risk category because of the new mask, with severity and position held at their published values; and
\item the mean mask IoU against the reference masks must be at least 0.85, with at least 90\% of masks above 0.8.
\end{enumerate}

\subsection{Stage 2: Segmentation}
Five lightweight SAM variants (MobileSAM, TinySAM, EfficientViT-SAM-L2 and -XL1, and SAM~2.1-small) were tested in one shared measurement loop. Each model has a thin adapter so that the prompts, the rule for choosing among SAM's candidate masks (highest predicted IoU) and the timing code are identical. We also tested a ``windowed'' setting in which each box is cropped with four times its size as context (at least 384 pixels) before segmentation, because small objects can benefit from a closer view.

Because all variants failed the criteria (Section~\ref{sec:seg-results}), we kept the published ViT-H model and changed only how it runs. The image encoder runs under fp16 automatic mixed precision~\cite{micikevicius2018mixed}, while the prompt encoder and mask decoder stay in fp32, and the model is compiled with \texttt{torch.compile}~\cite{ansel2024pytorch2}. Mixed precision keeps numerically sensitive operations such as normalisation and reductions in fp32; we verify the effect on the masks directly rather than assume it.

\subsection{Stage 3: Geometric and Positional Features}
For each mask, Stage~3 computes the features of the prior work~\cite{aburaddaha2026pothole}: area $A$ (mask pixels), perimeter $P$, circularity $4\pi A/P^2$, aspect ratio of the bounding rectangle, and solidity. It also assigns a distance band (near, medium or far, from the box's lower edge) and a lane zone (in the left or right wheel path, or at the left or right lane edge, from the box's left edge). The models keep the published solidity definition (mask pixel area over contour hull area) for consistency with the cloud system; the dataset also includes a contour-based solidity bounded by one.

\subsection{Stage 4: Severity Grading and Risk Scoring}
\label{sec:distill}
Severity has three classes that follow the prior work's rubric: low (about 2~cm deep or less), medium (2--5~cm) and high (more than 5~cm). These are visual severity grades aligned with maintenance thresholds, not calibrated depth measurements.

\paragraph{Local VLM candidates.} As a baseline we tested open-weight VLMs: Qwen3-VL-8B and Qwen3-VL-30B-A3B (8-bit floating point)~\cite{bai2025qwen3vl}. Both received the published severity prompt word for word, under JSON-schema decoding at temperature 0.1. Two input fixes were needed for a fair test. By default the model's image processor gave a typical $116 \times 121$ pothole crop only about 72 visual tokens, while leaving 7-megapixel scenes uncapped. We therefore upscaled crops to at least 336 pixels on the short side and capped scenes at 1536 pixels. The second fix, the order of the output fields, is reported in Section~\ref{sec:field-order}.

\paragraph{Distilled student ensemble.} We trained ConvNeXt-Tiny students~\cite{liu2022convnext} (28~M parameters, ImageNet initialisation) to predict the cloud model's severity label from the pothole crop. Data were split \emph{by scene} rather than by pothole, because potholes in one frame share lighting, camera pose and road surface; 15\% of scenes formed the validation set. Training used AdamW (learning rate $3\times10^{-4}$, weight decay $10^{-4}$, cosine schedule, 8 epochs, batch size 64), random resized crops, horizontal flips and colour jitter, and class-weighted cross-entropy to offset the class imbalance. We trained two variants: an image-only model, and a two-tower model that also receives a six-value geometry vector (normalised log-area, circularity, solidity, aspect ratio, distance band and an in-path flag) through a small multilayer perceptron. Each variant was trained with three seeds under fully deterministic settings (seeded data-loader workers and deterministic GPU kernels), so that the spread across training runs can be measured. The deployed grader is the majority vote of the six models, and the vote margin is passed to Stage~5 as a confidence value.

\paragraph{Risk score.} The composite risk score is unchanged from the published implementation:
\begin{equation}
R = 0.5\,S_d + 0.3\,S_a + 0.2\,S_p ,
\label{eq:risk}
\end{equation}
where $S_d$ is 20, 60 or 100 for low, medium or high severity; $S_a = \mathrm{clip}\big(100\,(\log_{10}A - 2)/2.5,\,0,\,100\big)$ rewards larger potholes on a log scale; and $S_p$ is 100 in a wheel path, 70 at a lane edge and 30 elsewhere. A pothole is critical if $R \ge 80$, high if $R \ge 60$, medium if $R \ge 40$ and low otherwise. The scene's reference urgency is the category of its highest-risk pothole.

\subsection{Stage 5: Driver Warning Generation}
\label{sec:stage5}
Qwen3-VL-8B~\cite{bai2025qwen3vl} writes each warning from the published multimodal prompt: a fixed rubric that defines the severity, distance and risk categories, the scene image, and the scene's assessment record from Stages~3--4. Decoding uses temperature 0.3 and a JSON schema with seven fields, in this order: \texttt{reasoning}, \texttt{maneuver\_type}, \texttt{warning\_message}, \texttt{maneuver\_suggestion}, \texttt{urgency\_level} (low, medium, high or critical), \texttt{affected\_lanes} and \texttt{recommended\_speed\_reduction}.

\paragraph{Closed maneuver vocabulary.} The pipeline sees only the ego lane. It cannot see whether the adjacent lane is free, so advice to change lanes is advice it cannot justify. Rather than forbid lane changes in the prompt, which the model can ignore, we removed them from the output space. The \texttt{maneuver\_type} field may take only one of six in-lane actions (Table~\ref{tab:maneuvers}), and the grammar enforces this at every decoding step. A lane change is then not discouraged but impossible to express in the structured field.

\begin{table}[H]
\caption{The closed maneuver vocabulary enforced by the decoding grammar.\label{tab:maneuvers}}
\newcolumntype{C}{>{\centering\arraybackslash}X}
\begin{tabularx}{\textwidth}{lX}
\toprule
\textbf{Maneuver type} & \textbf{Meaning} \\
\midrule
\texttt{maintain\_lane\_and\_speed} & no action needed \\
\texttt{reduce\_speed} & slow down and stay centred in the lane \\
\texttt{shift\_within\_lane\_left} & small offset to the left, staying inside the lane \\
\texttt{shift\_within\_lane\_right} & small offset to the right, staying inside the lane \\
\texttt{shift\_within\_lane\_and\_reduce\_speed} & small in-lane offset combined with slowing down \\
\texttt{brake\_firmly} & severe hazard with no room to move sideways \\
\bottomrule
\end{tabularx}
\end{table}

\paragraph{Serving.} The model is served with vLLM~\cite{kwon2023vllm} using the XGrammar backend~\cite{dong2024xgrammar} for structured output. The prompt is ordered as [fixed rubric] $\rightarrow$ [image] $\rightarrow$ [scene-specific data], so that the roughly 1500-token rubric is shared across all scenes and served from the prefix cache. Prefix caches match only identical leading tokens, so the shared content must come first. The schema contains only structural constraints, for the reason given in Section~\ref{sec:throughput}.

\subsection{Evaluation Protocol}
\label{sec:protocol}

\paragraph{Severity.} Every severity grader is scored with Cohen's $\kappa$~\cite{cohen1960} and raw agreement against the 38 consensus labels. The cloud model is scored from its released labels, the same labels the cloud pipeline produced and the students learned from; on this set they reach $\kappa = 0.321$ (55.3\% agreement), and the mean pairwise $\kappa$ between human annotators is 0.453. The prior work also reports $\kappa = 0.479$ for the cloud model on this panel; that value comes from a separate validation run whose labels differ from the released labels on 8 of the 38 items~\cite{aburaddaha2026pothole}. We use the released labels throughout. For the distilled students we report the spread across independently trained runs, not only a bootstrap over test items (Section~\ref{sec:severity}).

\paragraph{Warnings, automatic measures.} Over all 1119 scenes we measure: (i) \emph{rubric consistency}, the share of scenes whose generated urgency equals the reference urgency from Equation~(\ref{eq:risk}), with bootstrap confidence intervals; and (ii) the \emph{lane-departure rate}, the share of warnings that advise leaving the lane, compared between systems with McNemar's test~\cite{mcnemar1947}. Rubric consistency measures whether the generator follows the urgency implied by its own inputs.

\paragraph{Human study.} Three raters compared the local and cloud warnings for the same 50 scenes. The scenes were a fresh random sample from the full corpus, stratified by the local system's urgency in proportion to its distribution (9 critical, 14 high, 23 medium and 4 low); only 2 of them had appeared in an earlier pilot round (Appendix~\ref{app:round1}). For each scene, a spreadsheet showed the scene image and two warnings, labelled A and B, each with its message and maneuver advice. Which system was A was decided at random for each scene (the local system was A in 23 scenes), and the key was kept from the raters. Each rater scored each warning for \emph{clarity} (clear, well structured, easy to understand) and \emph{safety appropriateness} (suitable for a safety-critical driving decision) on 1--5 scales, and then chose the warning they would prefer while driving, or a tie. Raters were asked to work alone and not to discuss the task.

The three raters do not know one another and worked independently. Two are researchers, one in the United States and one in Germany; the third is an everyday driver with no research background, included so that the panel reflects the people who would actually receive the warnings. All three are adult volunteers. They received written instructions and three filled-in training examples on fictional scenes that were not part of the study. All 300 ratings (50 scenes $\times$ 2 systems $\times$ 3 raters) were returned complete.

\paragraph{Statistical analysis.} The unit of analysis is the scene. The three ratings of one scene are not independent observations of a system, so we first average them within each scene and system and then compare systems with the two-sided Wilcoxon signed-rank test~\cite{wilcoxon1945} on the 50 paired scene means, with a paired $t$-test as a check. Confidence intervals come from 10{,}000 bootstrap resamples of scenes~\cite{efron1979bootstrap}, and effect size is reported as $d_z$ (mean paired difference divided by its standard deviation). Preferences are tested with a two-sided binomial test that excludes ties. Rater agreement is reported as the two-way random-effects intraclass correlation for a single rater, ICC(2,1), and for the mean of three raters, ICC(2,3)~\cite{shrout1979icc,koo2016icc}, together with ordinal Krippendorff's $\alpha$~\cite{hayes2007kripp}, and Fleiss' $\kappa$~\cite{fleiss1971} for preferences.

\paragraph{Automated judge.} For comparison we also used Gemini~2.5~Flash as an automated judge, following common practice~\cite{zheng2023judging,wang2024notfair}. The judge saw the scene image and both warnings, scored each for clarity and safety on the same scales, and gave a preference, at temperature 0. Each pair was judged twice with the A/B order swapped, and only preferences that agreed across both orders were counted. The judge scored 150 scenes drawn at random from the corpus. It shares a model family with the cloud generator, a known source of bias~\cite{panickssery2024selfpref}.

\subsection{Hardware and Software}
All experiments ran on a workstation with two NVIDIA RTX 6000 Ada GPUs (48~GB each), using one GPU at a time, with vLLM 0.11.0, PyTorch 2.8.0 (CUDA 12.8) and Transformers 4.57.1. The 30B model was served from an 8-bit floating-point checkpoint so that it fits on a single GPU.

\section{Results}
\label{sec:results}

\subsection{Where the Time Goes}
Profiling the cloud pipeline component by component shows that Stage~2+3 inference takes 0.370~s per scene (image encoder 0.318~s, mask decoder 0.017~s, features 0.008~s); the rest of the published Stage-2 time is spent writing visualisations to disk, which a deployed system does not need. The two cloud calls take about 13.3~s (severity) and 3.5~s (warning), so about 98\% of the per-scene time is spent waiting for the cloud API. Replacing these two stages is therefore where almost all of the speed-up lies.

\subsection{Segmentation}
\label{sec:seg-results}
\begin{figure}[H]
\centering
\includegraphics[width=0.78\textwidth]{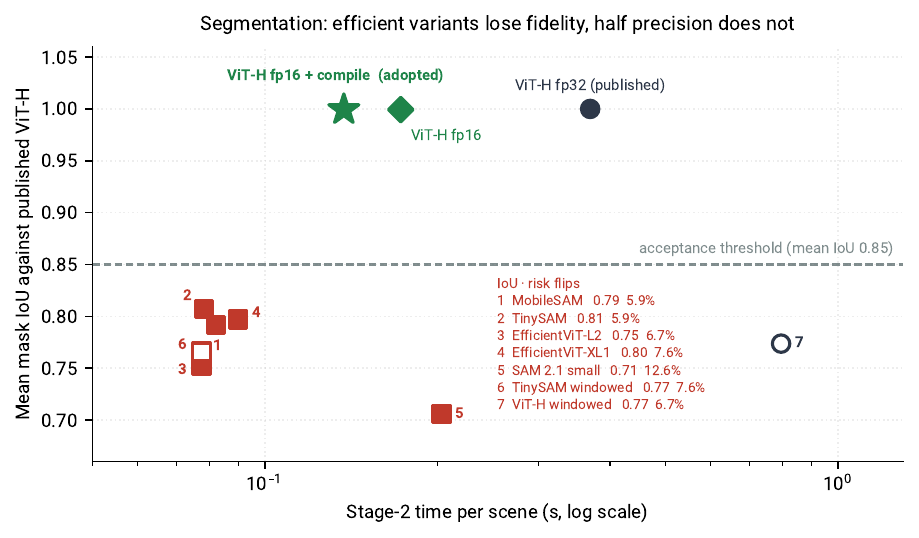}
\caption{Stage-2 candidates: mask fidelity against the published ViT-H masks versus time per scene (50 scenes, 119 potholes). Half precision with compilation keeps the masks and is 2.7$\times$ faster than the published model; the lightweight variants fall below the acceptance threshold. Hollow markers are windowed settings.\label{fig:seg}}
\end{figure}

Table~\ref{tab:seg} and Figure~\ref{fig:seg} compare the candidates. The original ViT-H with an fp16 image encoder and compilation reproduces the published masks (IoU 0.9993 against fp32, area error 0.05\%, no risk-category change over 119 potholes) at 0.137~s per scene, 2.7$\times$ faster than the published model. None of the lightweight variants meets the criteria. Even EfficientViT-SAM-XL1, which matches ViT-H on COCO and LVIS~\cite{zhang2024efficientvitsam}, reaches a mean IoU of 0.797 on potholes, and 7.6\% of potholes change risk category from mask-area error alone. Cropping each box with extra context (``windowed'') does not help: it lowers the IoU of every encoder, including ViT-H (0.774), which shows that the surrounding road context matters for these targets.

\begin{table}[H]
\caption{Stage-2 candidates against the published ViT-H masks (50 scenes, 119 potholes, annotated box prompts). ``$>$0.9'' is the share of masks with IoU above 0.9. Risk changes are computed with severity and position held at their published values. Acceptance requires mean IoU $\ge$ 0.85 and risk changes $\le$ 2\%.\label{tab:seg}}
\newcolumntype{C}{>{\centering\arraybackslash}X}
\begin{tabularx}{\textwidth}{lCCCCC}
\toprule
\textbf{Model} & \textbf{s/scene} & \textbf{Mean IoU} & \textbf{$>$0.9} & \textbf{Area error} & \textbf{Risk change} \\
\midrule
ViT-H (published, fp32) & 0.369 & 1.000 & 100\% & 0\% & 0\% \\
MobileSAM~\cite{zhang2023mobilesam} & 0.082 & 0.791 & 16\% & 19.7\% & 5.9\% \\
TinySAM~\cite{shu2025tinysam} & 0.078 & 0.807 & 20\% & 17.0\% & 5.9\% \\
EfficientViT-SAM-L2~\cite{zhang2024efficientvitsam} & 0.077 & 0.752 & 8\% & 26.6\% & 6.7\% \\
EfficientViT-SAM-XL1~\cite{zhang2024efficientvitsam} & 0.090 & 0.797 & 16\% & 21.3\% & 7.6\% \\
SAM~2.1-small~\cite{ravi2024sam2} & 0.203 & 0.706 & 5\% & 45.9\% & 12.6\% \\
\midrule
TinySAM, windowed & 0.078 & 0.766 & 19\% & 16.7\% & 7.6\% \\
ViT-H, windowed (control) & 0.796 & 0.774 & 21\% & 17.9\% & 6.7\% \\
\midrule
ViT-H, fp16 encoder & 0.173 & 0.9993 & 100\% & 0.05\% & 0\% \\
\textbf{ViT-H, fp16 encoder + compile} & \textbf{0.137} & \textbf{0.9993} & \textbf{100\%} & \textbf{0.05\%} & \textbf{0\%} \\
\bottomrule
\end{tabularx}
\end{table}

Because ViT-H masks are the reference above, we also checked them against human masks. On PothRGBD's hand-drawn masks, the box-prompted ViT-H reaches a mean IoU of 0.780 (median 0.809), with 96.0\% of masks above 0.5. SAM's own predicted IoU (mean 0.953) overestimates this and correlates only moderately with the measured IoU ($r = 0.569$), so we rely on measured agreement rather than on the model's self-reported confidence.

\subsection{Severity Grading}
\label{sec:severity}
Table~\ref{tab:depth} lists every grader on the 38-item consensus set. The distilled ensemble reaches $\kappa = 0.268 \pm 0.056$ at 28~ms per pothole, close to the cloud model's 0.321 and well above the open-weight VLMs. With the published prompt unchanged, Qwen3-VL-8B gave every pothole the majority class ($\kappa = 0.007$); the input fixes raised this to $\kappa = 0.152$, and the larger 30B model reached 0.103. The $\pm$ term of the ensemble is the standard deviation across three ensembles trained independently from scratch (18 models in all). Single students average $\kappa = 0.265 \pm 0.049$, so the ensemble's main benefit is a more stable prediction and a vote margin that serves as a confidence value. We report this spread across training runs, rather than only a bootstrap over test items, because for a small fine-tuned model the variation between runs is the larger of the two.

\begin{table}[H]
\caption{Severity graders on the 38-item human consensus set. All $\kappa$ values are computed on the same labels. For the distilled rows, $\pm$ is the standard deviation across independent training runs (3 ensembles, 18 models), not a bootstrap over test items.\label{tab:depth}}
\newcolumntype{C}{>{\centering\arraybackslash}X}
\small
\begin{tabularx}{\textwidth}{lCCC}
\toprule
\textbf{Grader} & \textbf{Cohen's $\kappa$} & \textbf{Agreement} & \textbf{Time per pothole} \\
\midrule
Mean pairwise human & 0.453 & -- & -- \\
Cloud VLM (Gemini 2.5 Pro, released labels) & 0.321 & 55.3\% & 4.8~s (API) \\
\midrule
Qwen3-VL-8B, published prompt & 0.007 & 39.5\% & $\sim$12~s \\
\quad + input resizing & 0.093 & 44.7\% & $\sim$12~s \\
\quad + resizing, reasoning first & 0.152 & 44.7\% & $\sim$14~s \\
Qwen3-VL-30B-A3B (FP8) & 0.103 & 39.5\% & $\sim$13~s \\
\midrule
Distilled student, single (mean of 18) & $0.265 \pm 0.049$ & -- & 4.4~ms \\
\textbf{Distilled 6-model ensemble} & $\mathbf{0.268 \pm 0.056}$ & -- & \textbf{28~ms} \\
\bottomrule
\end{tabularx}
\end{table}

\paragraph{Agreement with the teacher.} Over the full corpus the student follows its teacher closely: 74.0\% raw agreement, $\kappa = 0.372$ (95\% CI [0.34, 0.41]) and quadratic-weighted $\kappa = 0.502$, with class counts of 2250/578/264 (low/medium/high) against the teacher's 2316/603/173.

\subsection{Field Order in Constrained Decoding}
\label{sec:field-order}
The order of fields in the JSON schema strongly affects the open-weight VLM. When the class field comes first, the model must commit to an answer within a few tokens; under uncertainty it falls back to the majority class, and the reasoning written afterwards only justifies that choice. With resizing alone, predictions stayed concentrated in one class ($\kappa = 0.093$). Putting the reasoning field before the class field raised $\kappa$ to 0.152 and the model used all three classes. Evidence fields should therefore come before conclusion fields, and the warning schema of Section~\ref{sec:stage5} follows this rule.

\subsection{Severity Under a Different Camera Geometry}
\label{sec:domain}
\begin{figure}[H]
\centering
\includegraphics[width=0.8\textwidth]{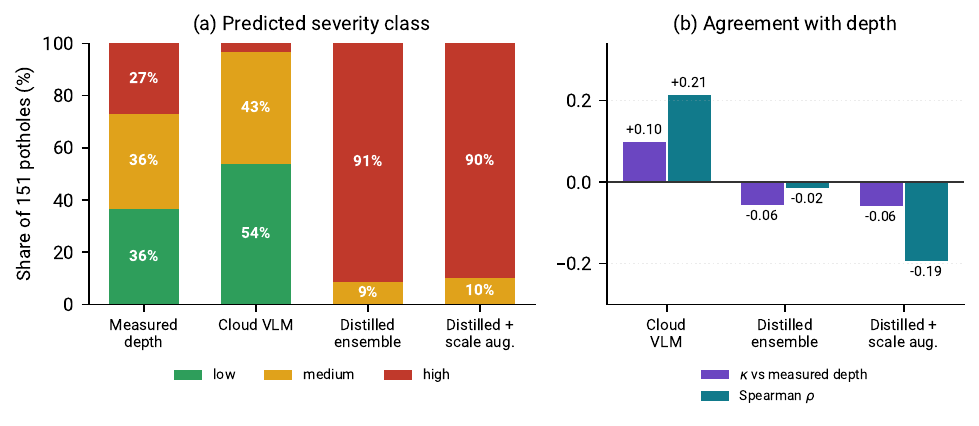}
\caption{Severity on 151 potholes from an independent close-range RGB-D dataset. (\textbf{a}) Predicted class distribution: both distilled ensembles put about 90\% of potholes in the top class, whereas the cloud VLM spreads its predictions. (\textbf{b}) Agreement with sensor-measured depth.\label{fig:domain}}
\end{figure}

To define where the distilled grader can be used, we applied it to a class-balanced sample of 151 potholes from PothRGBD, a close-range RGB-D dataset, and scored it against depth measured by the sensor. The training crops are distant dashcam defects with a median mask area of 3606 pixels, whereas PothRGBD potholes are photographed from about a metre away with a median of 68{,}560 pixels. On these images the ensemble assigns the highest class to 138 of the 151 potholes ($\kappa = -0.057$; Figure~\ref{fig:domain}), because large, detailed crops are what it learned to associate with severe potholes. The cloud VLM keeps a weak ordering on the same sample ($\kappa = 0.098$; Spearman $\rho = 0.214$, $p = 0.009$). Strong scale and crop augmentation during training did not change the student's behaviour ($\kappa = -0.059$). The distilled grader is therefore a camera-specific component: it is accurate for the dashcam geometry it was trained on, and a new camera setup calls for re-distillation from a one-off teacher labelling of images from that camera (about \$0.08 per scene with the cloud model).

\subsection{Warning Generation: Rubric Consistency and In-Lane Advice}
\label{sec:warnings}
Over all 1119 scenes, the urgency written by the constrained local generator matches the reference urgency in 94.8\% of scenes (95\% CI [93.5, 96.1]). Against the cloud system on the same scenes, urgency agreement is 68.4\% ($\kappa = 0.514$). The local system issues more high-urgency and fewer medium-urgency warnings (Stuart--Maxwell $\chi^2 = 73.5$, df $= 3$, $p < 10^{-15}$), which follows from its severity inputs.

The closed maneuver vocabulary is what makes the local warnings safe to act on. Without it, the local generator advised \emph{leaving the lane} in 542 of 1119 scenes (48.4\%), against 4 (0.4\%) for the cloud system (Figure~\ref{fig:safety}a), even though neither system can see the adjacent lane. Table~\ref{tab:advisory} reports the controlled comparison: both local versions use identical assessment records and decoding settings and differ only in the maneuver constraint. The closed vocabulary cuts lane-departure advice from 48.4\% to 6.1\% of scenes, correcting 505 scenes while introducing 31 (McNemar $p = 1.9\times10^{-111}$), and it leaves rubric consistency unchanged (94.8\% vs.\ 95.5\%). The remaining 6.1\% are scenes whose free-text explanation still mentions an adjacent lane, although the structured maneuver cannot. The constrained generator chooses \texttt{reduce\_speed} in 1089 scenes, \texttt{maintain\_lane\_and\_speed} in 16 and \texttt{brake\_firmly} in 14: a conservative default that the human raters rewarded (Section~\ref{sec:human}).

\begin{table}[H]
\caption{Warning quality on all 1119 scenes. Both local versions use the same assessment records and decoding settings and differ only in whether the grammar restricts the maneuver field. Judge scores are from the automated judge on 150 scenes (Section~\ref{sec:judge}); the cloud scores are those from the pairing with the constrained version. Judge preferences count only verdicts that agree across both presentation orders.\label{tab:advisory}}
\newcolumntype{C}{>{\centering\arraybackslash}X}
\small
\begin{tabularx}{\textwidth}{p{4.3cm}CCC}
\toprule
 & \textbf{Cloud (prior work)} & \textbf{Local, unconstrained} & \textbf{Local, constrained} \\
\midrule
Rubric consistency & -- & 95.5\% [94.3, 96.7] & 94.8\% [93.5, 96.1] \\
\textbf{Lane-departure advice} & \textbf{0.4\%} & \textbf{48.4\%} & \textbf{6.1\%} \\
Judge clarity (1--5) & 4.60 & 3.29 & 3.78 \\
Judge safety (1--5) & 4.20 & 2.43 & 2.97 \\
Judge preference (cloud : local) & -- & 110 : 14 & 103 : 17 \\
\bottomrule
\end{tabularx}
\end{table}

\subsection{Throughput}
\label{sec:throughput}
The schema design also determines speed. With \texttt{maxLength} limits on the string fields, generation took about 30~s per scene (6.6 tokens per second), because a character-length limit must be tracked inside the per-step token mask, and its cost grows quickly when tokens span several characters. Keeping all structural constraints but removing the length limits raised throughput to 316 tokens per second and cut generation to 0.71~s per scene, a 44$\times$ improvement, with no loss in rubric consistency. Multimodal serving also needs spare GPU memory for the bursts of vision-encoder activations during image prefill, so the memory share given to vLLM should not be set too high.

\subsection{Human Evaluation of the Warnings}
\label{sec:human}
\begin{figure}[H]
\centering
\includegraphics[width=0.95\textwidth]{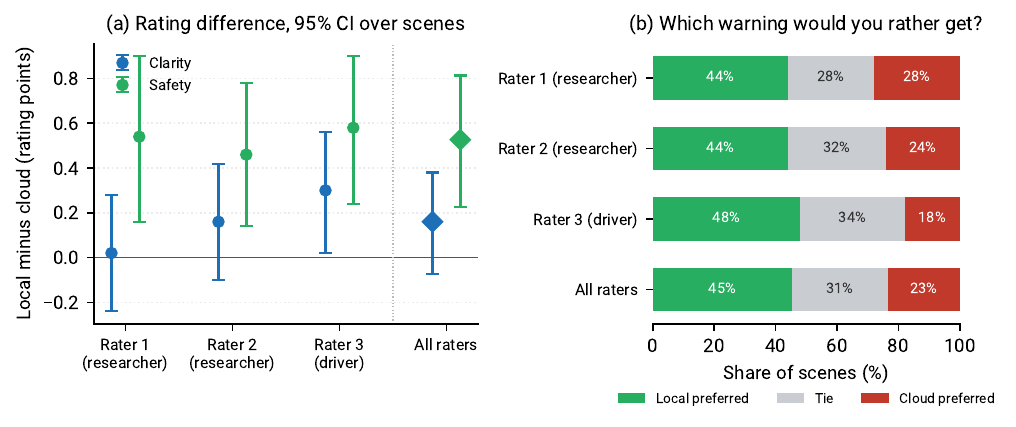}
\caption{Human study, rater by rater. (\textbf{a}) Local-minus-cloud difference in mean rating with 95\% bootstrap intervals over the 50 scenes; diamonds pool the three raters. Every rater scored the local warnings safer. (\textbf{b}) Preference split for each rater.\label{fig:human}}
\end{figure}

Table~\ref{tab:human} and Figure~\ref{fig:human} give the outcome of the blind study on the reported (constrained) local system. Raters scored the local warnings \textbf{safer} than the cloud warnings by 0.53 points on the 5-point scale (4.18 vs.\ 3.65; 95\% CI [0.23, 0.81]; Wilcoxon $p = 0.001$; $d_z = 0.50$), with the local warning rated safer in 33 scenes, the cloud warning in 9 and a tie in 8. Clarity was similar (4.18 vs.\ 4.02; difference $+0.16$, 95\% CI [$-0.06$, 0.37]; $p = 0.14$). Across all 150 judgements, raters preferred the local warning 68 times and the cloud warning 35 times, with 47 ties (binomial $p = 0.0015$, ties excluded). By scene-level majority the split was 20 local, 12 cloud, 16 tie and 2 with no majority.

\begin{table}[H]
\caption{Blind human comparison of the driver warnings (50 scenes, 3 independent raters, 1--5 scales). Means are over all 150 ratings per system. Differences are local minus cloud, tested on the 50 per-scene means with the Wilcoxon signed-rank test; intervals are 95\% bootstrap intervals over scenes.\label{tab:human}}
\newcolumntype{C}{>{\centering\arraybackslash}X}
\small
\begin{tabularx}{\textwidth}{lCCCCC}
\toprule
 & \textbf{Cloud} & \textbf{Local} & \textbf{Difference} & \textbf{95\% CI} & \textbf{$p$} \\
\midrule
Safety appropriateness & 3.65 $\pm$ 0.86 & \textbf{4.18 $\pm$ 0.81} & $\mathbf{+0.53}$ & [0.23, 0.81] & 0.001 \\
Clarity & 4.02 $\pm$ 0.76 & 4.18 $\pm$ 0.75 & $+0.16$ & [$-0.06$, 0.37] & 0.14 \\
\midrule
Preferred (of 150) & 35 & \textbf{68} & 47 ties & -- & 0.0015 \\
\bottomrule
\end{tabularx}
\end{table}

\paragraph{Consistency across raters.} The result does not rest on one rater. Each of the three raters, on their own, scored the local warnings safer, by 0.54, 0.46 and 0.58 points (Wilcoxon $p = 0.010$, 0.012 and 0.002), and each preferred the local warning more often than the cloud warning (22:14, 22:12 and 24:9). The everyday driver showed the largest preference for the local warnings. Agreement between raters was moderate for a single rater and good for their mean: ICC(2,1) was 0.60 for safety and 0.53 for clarity, and ICC(2,3) was 0.82 and 0.77~\cite{koo2016icc}; ordinal Krippendorff's $\alpha$ was 0.60 and 0.52. Agreement on the preference itself was fair (Fleiss' $\kappa = 0.25$). These values are typical of independent raters judging short texts, and they show that the panel shared a view without being uniform. The safety advantage held in every urgency stratum (critical: 4.07 vs.\ 3.67; high: 4.26 vs.\ 3.64; medium: 4.23 vs.\ 3.65; low: 3.83 vs.\ 3.67, the last with only 4 scenes).

\begin{figure}[!tbp]
\begin{adjustwidth}{-\extralength}{0cm}
\centering
\includegraphics[width=\linewidth]{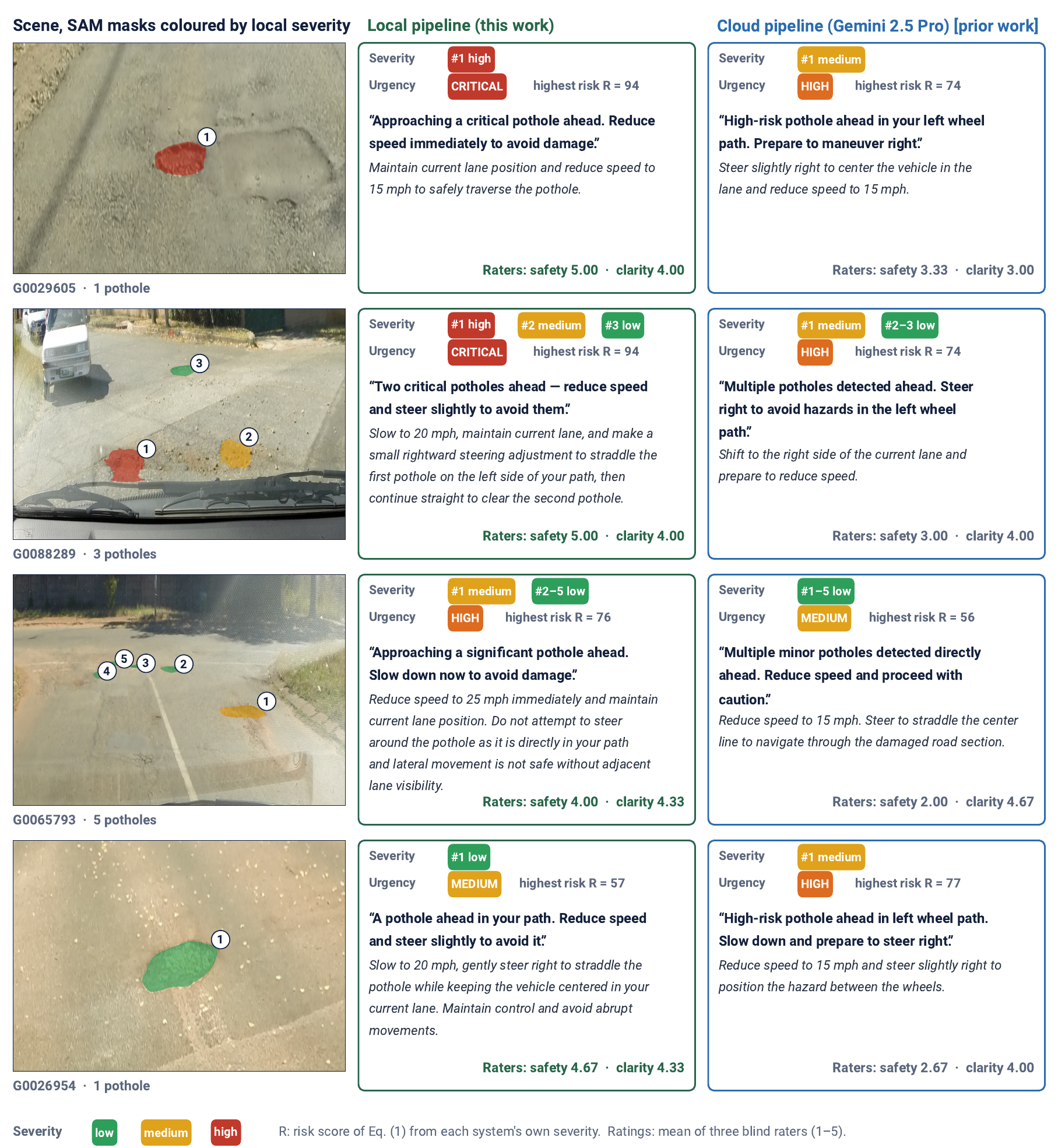}
\end{adjustwidth}
\caption{Qualitative comparison on four scenes from the blind human study. Left: the perspective-corrected scene with the SAM masks, coloured by the local severity grade and numbered. Middle and right: the per-pothole severity, the highest risk score $R$ (Equation~(\ref{eq:risk}), computed from each system's own severity), the urgency and the warning (message in bold, maneuver advice in italics) of the local and the cloud pipeline, exactly as the raters saw them. The ratings are the mean safety and clarity scores of the three raters for that scene (1--5).\label{fig:qualitative}}
\end{figure}

\paragraph{Qualitative results.} Figure~\ref{fig:qualitative} shows four scenes from the study with the full output of both pipelines: the masks, each system's per-pothole severity and risk, the urgency, and the warning as the raters saw it. The examples illustrate what the raters rewarded. Every local warning asked the driver to slow down, and 98\% gave a concrete target speed, against 78\% and 64\% for the cloud warnings. The local system also kept the vehicle in its lane, whereas 94\% of the cloud warnings asked the driver to steer or shift (72\% of the local ones), sometimes in ways that are hard to justify from a single camera, such as straddling the centre line (G0065793) or shifting across the lane before slowing down (G0088289). In the first two scenes the local grader also rated the most damaged pothole as high severity where the cloud model rated it medium, and its warning raised the urgency to critical accordingly. The cloud warnings scored higher in a few dense scenes in which several large potholes blocked the lane and its firmer ``brake immediately'' advice was preferred.

\subsection{The Automated Judge Versus Human Raters}
\label{sec:judge}
\begin{figure}[H]
\centering
\includegraphics[width=0.92\textwidth]{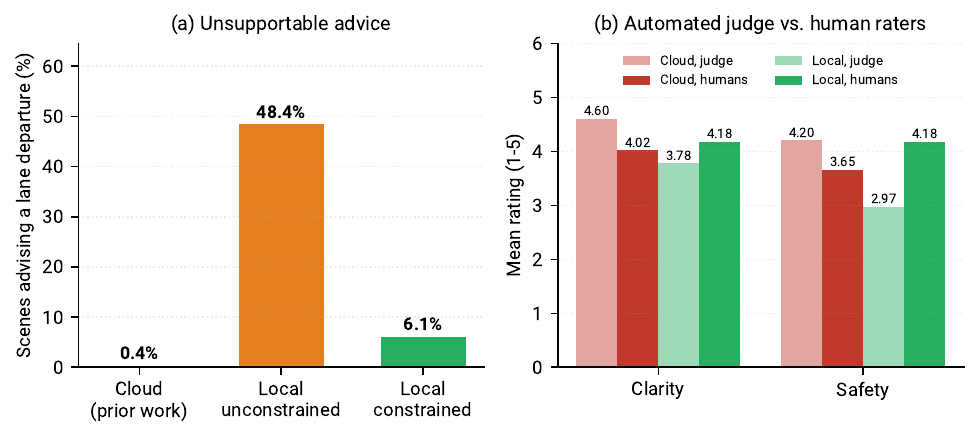}
\caption{(\textbf{a}) Warnings that advise leaving the lane. Neither system observes the adjacent lane; the closed maneuver vocabulary cuts it from 48.4\% to 6.1\% of scenes. (\textbf{b}) Mean clarity and safety ratings for the reported local system and the cloud system, from the automated judge (150 scenes) and from the three human raters (50 scenes). Humans rate the local warnings safer; the judge rates them less safe.\label{fig:safety}}
\end{figure}

The automated judge reached the opposite verdict (Figure~\ref{fig:safety}b). On the same pair of systems it scored the local warnings 1.23 points \emph{less} safe than the cloud warnings (2.97 vs.\ 4.20) and 0.82 points less clear (3.78 vs.\ 4.60), and it preferred the cloud warning in 103 of the 120 order-consistent pairs ($p < 10^{-15}$). Humans and judge thus disagree in direction, not only in size. Compared with the human means, the judge rated the local warnings 1.21 points lower on safety and the cloud warnings 0.55 points higher, an asymmetry consistent with the judge's family link to the cloud generator~\cite{panickssery2024selfpref}. The judge and the humans scored different random samples of the same corpus, so this is a comparison of means rather than of individual items; the same direction of error appeared in a pilot round, where the judge rated the local warnings 1.52 points less safe than the human raters did (Appendix~\ref{app:round1}).

\subsection{End-to-End Cost and Latency}
\label{sec:e2e}
Table~\ref{tab:e2e} compares the complete systems. The local pipeline runs at 1.27~s per scene, a 16$\times$ end-to-end reduction, and regenerated the full corpus in 23.7 minutes at no marginal cost. These times come from the production run that generated the released corpus, which shared the GPU with another process; on an unshared GPU, warning generation takes 0.71~s instead of 1.08~s per scene. Because the output is valid JSON by construction, the cloud system's fallback path for unparseable responses is no longer needed.

\begin{table}[H]
\caption{End-to-end comparison on identical scenes and prompts. Local timings are from the production run that generated the released corpus, measured while sharing the GPU with another process. The cloud corpus time is the measured wall-clock time of the prior run, including API queueing and retries (26.5~s per scene)~\cite{aburaddaha2026pothole}.\label{tab:e2e}}
\newcolumntype{C}{>{\centering\arraybackslash}X}
\begin{tabularx}{\textwidth}{lCC}
\toprule
 & \textbf{Cloud pipeline~\cite{aburaddaha2026pothole}} & \textbf{This work (local)} \\
\midrule
Stage 2: segmentation & 0.369~s & 0.137~s \\
Stage 3: features & $\sim$0.05~s & 0.009~s \\
Stage 4: severity and risk & $\sim$13.3~s (API) & 0.030~s \\
Stage 5: warning & $\sim$3.5~s (API) & 1.080~s \\
\midrule
\textbf{Total per scene} & $\sim$\textbf{20.1~s} & \textbf{1.27~s} \\
Marginal cost per scene & $\sim$\$0.105 & \$0 \\
Full corpus (1119 scenes) & 8.2~h and \$117.50 & 23.7~min \\
Network access & 2 API calls per scene & none \\
Unparseable outputs & possible (fallback path) & none (grammar) \\
\bottomrule
\end{tabularx}
\end{table}

\section{Discussion}
\label{sec:discussion}

\subsection{Severity Grading: Accuracy, Speed and Scope}
\begin{figure}[H]
\centering
\includegraphics[width=0.72\textwidth]{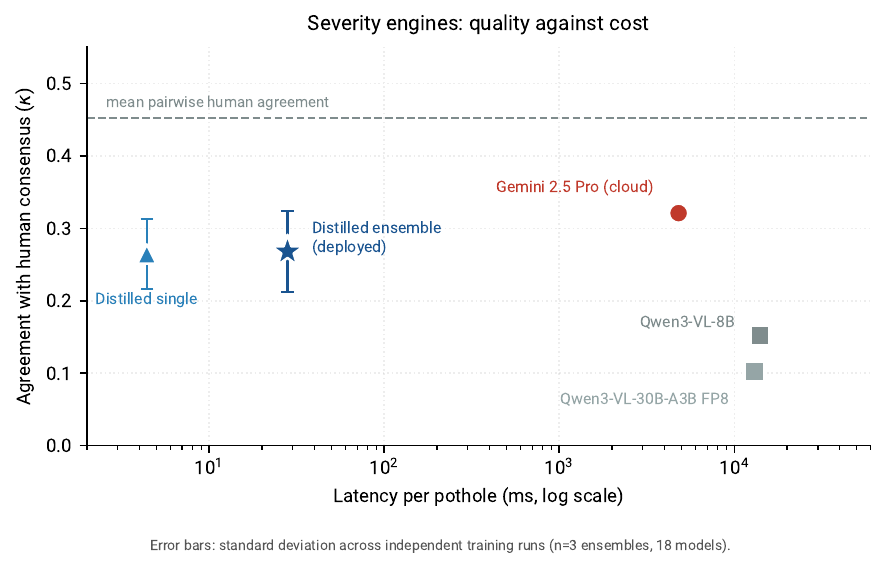}
\caption{Severity graders on the quality--cost plane (log time axis). Error bars are the standard deviation across three independent ensembles (18 models). The distilled ensemble is 170$\times$ faster than the cloud model at $\Delta\kappa = 0.053$, and well above the open-weight local VLMs at a thousandth of their latency.\label{fig:frontier}}
\end{figure}

Within its camera setup, the distilled ensemble agrees with human consensus 2.6 times better than a 30B open-weight VLM ($\kappa$ 0.268 vs.\ 0.103) at about a thousandth of the latency. Figure~\ref{fig:frontier} places the severity graders on a quality--cost plane. Against the cloud model, the local grader gives up $\Delta\kappa = 0.053$, which lies within its own spread across training runs ($\pm 0.056$) and is not significant on 38 items, in exchange for a 170$\times$ reduction in latency per pothole and no marginal cost. The out-of-domain test (Section~\ref{sec:domain}) defines the grader's scope: it is valid for the camera geometry it was distilled for, and a new camera setup is handled by re-distilling it once from teacher labels for that camera. Applications that want the cloud model's grading can also keep it for Stage~4 inside an otherwise local pipeline, at about \$0.080 per scene.

\subsection{Why Raters Preferred the Local Warnings}
The local generator is far smaller than the cloud model, yet its warnings were rated safer. We see three reasons. First, the local warnings are consistent: they always ask the driver to slow down and almost always give a concrete target speed, which raters found reassuring and easy to act on. Second, the closed vocabulary removes the riskiest class of advice, and the rubric is applied with high fidelity (94.8\% rubric consistency). Third, the cloud model, a strong general-purpose system, sometimes chose clever but risky maneuvers, such as steering to straddle the centre line (Figure~\ref{fig:qualitative}, G0065793), or kept the current speed for a defect in the wheel path. The gain therefore comes from the design of the output space rather than from model size: constraining the warning to actions the sensors can justify lets a compact local model produce advice that people judge safer. The cloud model's firmer advice was still preferred in some scenes with several large potholes, which suggests that a stronger \texttt{brake\_firmly} rule for blocked lanes would help further.

\subsection{What the Judge Result Means for Evaluation}
LLM judges are attractive because they are fast and cheap, and many papers now rely on them~\cite{zheng2023judging,chen2024mllmjudge}. Our judge was deployed as the literature recommends: blind to the source, at temperature zero, and with both presentation orders. It still reversed the human verdict on the question that matters most here, safety. Its family link to the cloud generator is one plausible cause~\cite{panickssery2024selfpref}; another is that a judge rewards fluent, varied language, whereas humans asked about safety reward caution and concreteness. A team that relied on the judge alone would have concluded that the local system was unsafe and abandoned it. For safety-related text, we recommend at least a small human study on the same items before trusting judge scores.

\subsection{Practical Guidance}
The study suggests several rules for moving a VLM pipeline from the cloud to local hardware:
\begin{itemize}
\item Profile before optimising. Here, almost all of the per-scene time was spent in the two API calls.
\item Prefer a faster execution of a trusted model over a smaller model when the targets are small and low-contrast; benchmark parity of lightweight models did not transfer to potholes.
\item Train distilled models deterministically and report the spread across training runs, not only an interval over test items.
\item Constrain what the model can say, not only what it is asked to say. A closed maneuver vocabulary removed most unsupportable advice at no cost to rubric consistency.
\item In JSON schemas, put reasoning fields before answer fields, and avoid character-length limits, which can slow constrained decoding by more than an order of magnitude.
\item Put shared prompt content first so that the prefix cache can reuse it, and leave extra GPU memory for image encoders.
\end{itemize}

\subsection{Limitations}
\label{sec:limitations}
\textbf{Severity grades.} Agreement with humans is moderate for both graders ($\kappa$ 0.268 for the distilled grader and 0.321 for the cloud model, against 0.453 between humans) on a 38-item panel; an independent, prediction-free panel in the prior work gave the cloud model $\kappa = 0.139$ against 0.605 between annotators~\cite{aburaddaha2026pothole}. The classes are visual grades, not calibrated depth, and the distilled grader is specific to the dashcam geometry it was trained on (Section~\ref{sec:domain}).

\textbf{Human study.} Fifty scenes and three raters bound the precision of the comparison; the safety difference is significant, and on clarity we claim parity. The ratings measure perceived appropriateness, not driving outcomes. Most of the 50 scenes were in the training split of some of the distilled students, so the study evaluates the warning stage; the two systems still received different severity inputs for about a quarter of potholes. The raters were volunteers known to the first author, though not to one another, and were blind to which system wrote each warning.

\textbf{Scope.} Annotated boxes stand in for an upstream detector, the main evaluation uses one dataset from one region, and all timings are on a workstation GPU. With the closed vocabulary, 6.1\% of warnings still mention an adjacent lane in their free text. Hosted cloud models also drift: repeating the cloud severity labelling three times gave identical classes for 91.3\% of potholes (Fleiss' $\kappa = 0.815$), so cloud results are reproducible only to within a few percent.

\section{Conclusions}
\label{sec:conclusions}
A cloud-dependent pothole assessment and warning pipeline can run entirely on local hardware. Segmentation runs 2.7$\times$ faster with unchanged downstream outputs; severity grading runs about 170$\times$ faster, within $\kappa$ 0.053 of the cloud model; and a constrained 8B model writes warnings that blind human raters scored safer than the cloud system's and equally clear. Together these make the pipeline 16$\times$ faster, free at the margin and independent of a network connection.

The study also gives practical guidance for such migrations. Execution-level acceleration of a trusted segmenter preserved accuracy where lightweight substitutes did not; a distilled grader is fast and accurate for its own camera geometry; field order and length limits in a constrained schema decide both the quality and the speed of generation; and safety-related text should be judged by humans, because an automated judge reversed their verdict. Future work will deploy the pipeline on embedded vehicle hardware, extend the closed vocabulary to the free-text explanation, add a detector in front of Stage~2, and use the ENPAD-A action labels to train vision--language--action models.

\vspace{6pt}

\authorcontributions{Conceptualization, A.A. and S.A.R.; methodology, A.A.; software, A.A.; validation, A.A.; formal analysis, A.A.; investigation, A.A.; data curation, A.A.; writing---original draft preparation, A.A.; writing---review and editing, A.A. and S.A.R.; visualization, A.A.; supervision, S.A.R.; project administration, S.A.R. All authors have read and agreed to the published version of the manuscript.}

\funding{This research received no external funding.}

\institutionalreview{Ethical review and approval were not sought for this study. The human evaluation asked three adult volunteers to rate machine-generated warning texts for public road images. The study collected information about the warning systems, not about the raters: no personal, health or identifying information about the raters was recorded, the task involved no intervention and no risk beyond ordinary computer work, and the ratings are reported only in aggregate and under anonymous labels.}

\informedconsent{All raters took part voluntarily, were told the purpose of the task, and agreed that their anonymous ratings could be used and reported in this study.}

\dataavailability{The code for both pipelines, the analysis scripts and a 30-scene sample of the ENPAD-A dataset are available at \url{https://github.com/malekaburaddaha/Pothole_Assessment_Pipeline}; the six trained severity models are attached to its release v2.0-paper2 (\url{https://github.com/malekaburaddaha/Pothole_Assessment_Pipeline/releases/tag/v2.0-paper2}). The full ENPAD-A dataset (1119 scenes and 3097 potholes, about 1.7~GB), the anonymised human ratings and all generated warnings and assessment records are available from the corresponding author on reasonable request. The underlying images remain subject to the licence of the original dataset~\cite{nienaber2015}. Model identifiers and versions are listed in the repository. The serving stack is pinned (vLLM 0.11.0, PyTorch 2.8.0, Transformers 4.57.1) because the pipeline proved sensitive to version changes.}

\acknowledgments{The authors thank the three volunteer raters for their careful and independent work. This work was carried out with the resources of the authors' laboratory at the University of Michigan--Dearborn. During the preparation of this manuscript, the authors used an AI assistant (Claude, Anthropic) to help build and modify some of the figures and to help draft and edit parts of the text. The authors reviewed and edited all output and take full responsibility for the content of this publication.}

\conflictsofinterest{The authors declare no conflicts of interest.}

\abbreviations{Abbreviations}{
The following abbreviations are used in this manuscript:
\\

\noindent
\begin{tabular}{@{}ll}
API & Application programming interface \\
CI & Confidence interval \\
FP8 & 8-bit floating point \\
fp16, fp32 & 16-bit and 32-bit floating point \\
ICC & Intraclass correlation coefficient \\
IoU & Intersection over union \\
LLM & Large language model \\
RGB-D & Colour image with depth \\
SAM & Segment Anything Model \\
ViT & Vision Transformer \\
VLM & Vision--language model
\end{tabular}
}

\appendixtitles{yes}
\appendixstart
\appendix
\section{Pilot Human Round on the Unconstrained Generator}
\label{app:round1}
Before the maneuver vocabulary was closed, a pilot round with the same protocol rated the unconstrained local generator against the cloud system on 49 scenes, with three raters from the research team. The two systems tied on safety (3.95 vs.\ 3.95), the cloud warnings were rated slightly clearer (4.19 vs.\ 3.96), and the preference split was 46 cloud, 38 local and 63 ties. In the main study, on the constrained system and with independent raters, the local warnings were rated safer (4.18 vs.\ 3.65) and equally clear. Because the system, the scene sample and the raters all changed between the rounds, the difference should not be attributed to the vocabulary constraint alone, although it agrees with the automatic measures in Table~\ref{tab:advisory}, which all improved when the vocabulary was closed. In the pilot, too, the automated judge rated the local system's safety far below the human raters (by 1.52 points).

\begin{adjustwidth}{-\extralength}{0cm}
\reftitle{References}
\bibliography{references}

\begin{thebibliography}{999}

\bibitem[Koch and Brilakis(2011)]{koch2011pothole}
Koch, C.; Brilakis, I.
\newblock Pothole detection in asphalt pavement images.
\newblock {\em Advanced Engineering Informatics} {\bf 2011}, {\em
  25},~507--515.
\newblock {\url{https://doi.org/10.1016/j.aei.2011.01.002}}.

\bibitem[Ma et~al.(2022)Ma, Fan, Wang, Wu, Jiang, Xie, and Fan]{ma2022review}
Ma, N.; Fan, J.; Wang, W.; Wu, J.; Jiang, Y.; Xie, L.; Fan, R.
\newblock Computer vision for road imaging and pothole detection: a
  state-of-the-art review of systems and algorithms.
\newblock {\em Transportation Safety and Environment} {\bf 2022}, {\em 4}.
\newblock {\url{https://doi.org/10.1093/tse/tdac026}}.

\bibitem[Aburaddaha et~al.(2026)Aburaddaha, El~Shair, Al-Bustami, and
  Rawashdeh]{aburaddaha2026pothole}
Aburaddaha, A.; El~Shair, Z.A.; Al-Bustami, A.; Rawashdeh, S.A.
\newblock Pothole Risk Assessment with Geometrically Grounded Vision Language
  Models.
\newblock {\em arXiv preprint arXiv:2611.16061} {\bf 2026}.
\newblock Submitted to IEEE Access.

\bibitem[Kirillov et~al.(2023)Kirillov, Mintun, Ravi, Mao, Rolland, Gustafson,
  Xiao, Whitehead, Berg, Lo, Doll{\'a}r, and Girshick]{kirillov2023sam}
Kirillov, A.; Mintun, E.; Ravi, N.; Mao, H.; Rolland, C.; Gustafson, L.; Xiao,
  T.; Whitehead, S.; Berg, A.C.; Lo, W.Y.;  et~al.
\newblock Segment Anything.
\newblock In Proceedings of the 2023 IEEE/CVF International Conference on
  Computer Vision (ICCV). IEEE,  2023, pp. 3992--4003.
\newblock {\url{https://doi.org/10.1109/iccv51070.2023.00371}}.

\bibitem[Eriksson et~al.(2008)Eriksson, Girod, Hull, Newton, Madden, and
  Balakrishnan]{eriksson2008pothole}
Eriksson, J.; Girod, L.; Hull, B.; Newton, R.; Madden, S.; Balakrishnan, H.
\newblock The pothole patrol: using a mobile sensor network for road surface
  monitoring.
\newblock In Proceedings of the Proceedings of the 6th international conference
  on Mobile systems, applications, and services. ACM,  2008, pp. 29--39.
\newblock {\url{https://doi.org/10.1145/1378600.1378605}}.

\bibitem[Mednis et~al.(2011)Mednis, Strazdins, Zviedris, Kanonirs, and
  Selavo]{mednis2011realtime}
Mednis, A.; Strazdins, G.; Zviedris, R.; Kanonirs, G.; Selavo, L.
\newblock Real time pothole detection using Android smartphones with
  accelerometers.
\newblock In Proceedings of the 2011 International Conference on Distributed
  Computing in Sensor Systems and Workshops (DCOSS). IEEE,  2011, pp. 1--6.
\newblock {\url{https://doi.org/10.1109/dcoss.2011.5982206}}.

\bibitem[Nienaber et~al.(2015)Nienaber, Booysen, and Kroon]{nienaber2015}
Nienaber, S.; Booysen, M.J.; Kroon, R.S.
\newblock Detecting Potholes Using Simple Image Processing Techniques and
  Real-World Footage.
\newblock In Proceedings of the Proceedings of the 34th Southern African
  Transport Conference (SATC), Pretoria, South Africa,  2015.

\bibitem[Ouma and Hahn(2017)]{ouma2017fuzzy}
Ouma, Y.O.; Hahn, M.
\newblock Pothole detection on asphalt pavements from 2D-colour pothole images
  using fuzzy c -means clustering and morphological reconstruction.
\newblock {\em Automation in Construction} {\bf 2017}, {\em 83},~196--211.
\newblock {\url{https://doi.org/10.1016/j.autcon.2017.08.017}}.

\bibitem[Jo and Ryu(2015)]{jo2015blackbox}
Jo, Y.; Ryu, S.
\newblock Pothole Detection System Using a Black-box Camera.
\newblock {\em Sensors} {\bf 2015}, {\em 15},~29316--29331.
\newblock {\url{https://doi.org/10.3390/s151129316}}.

\bibitem[Zhang et~al.(2016)Zhang, Yang, Daniel~Zhang, and Zhu]{zhang2016crack}
Zhang, L.; Yang, F.; Daniel~Zhang, Y.; Zhu, Y.J.
\newblock Road crack detection using deep convolutional neural network.
\newblock In Proceedings of the 2016 IEEE International Conference on Image
  Processing (ICIP). IEEE,  2016, pp. 3708--3712.
\newblock {\url{https://doi.org/10.1109/icip.2016.7533052}}.

\bibitem[Maeda et~al.(2018)Maeda, Sekimoto, Seto, Kashiyama, and
  Omata]{maeda2018rdd}
Maeda, H.; Sekimoto, Y.; Seto, T.; Kashiyama, T.; Omata, H.
\newblock Road Damage Detection and Classification Using Deep Neural Networks
  with Smartphone Images.
\newblock {\em Computer-Aided Civil and Infrastructure Engineering} {\bf 2018},
  {\em 33},~1127--1141.
\newblock {\url{https://doi.org/10.1111/mice.12387}}.

\bibitem[Arya et~al.(2021)Arya, Maeda, Ghosh, Toshniwal, Mraz, Kashiyama, and
  Sekimoto]{arya2021multicountry}
Arya, D.; Maeda, H.; Ghosh, S.K.; Toshniwal, D.; Mraz, A.; Kashiyama, T.;
  Sekimoto, Y.
\newblock Deep learning-based road damage detection and classification for
  multiple countries.
\newblock {\em Automation in Construction} {\bf 2021}, {\em 132},~103935.
\newblock {\url{https://doi.org/10.1016/j.autcon.2021.103935}}.

\bibitem[Arya et~al.(2024)Arya, Maeda, Ghosh, Toshniwal, and
  Sekimoto]{arya2024rdd2022}
Arya, D.; Maeda, H.; Ghosh, S.K.; Toshniwal, D.; Sekimoto, Y.
\newblock RDD2022: A multi-national image dataset for automatic road damage
  detection.
\newblock {\em Geoscience Data Journal} {\bf 2024}.
\newblock {\url{https://doi.org/10.1002/gdj3.260}}.

\bibitem[Ahmed(2021)]{ahmed2021dilated}
Ahmed, K.R.
\newblock Smart Pothole Detection Using Deep Learning Based on Dilated
  Convolution.
\newblock {\em Sensors} {\bf 2021}, {\em 21},~8406.
\newblock {\url{https://doi.org/10.3390/s21248406}}.

\bibitem[Jog et~al.(2012)Jog, Koch, Golparvar-Fard, and
  Brilakis]{jog2012properties}
Jog, G.M.; Koch, C.; Golparvar-Fard, M.; Brilakis, I.
\newblock Pothole Properties Measurement through Visual 2D Recognition and 3D
  Reconstruction.
\newblock In Proceedings of the Computing in Civil Engineering (2012). American
  Society of Civil Engineers,  2012, pp. 553--560.
\newblock {\url{https://doi.org/10.1061/9780784412343.0070}}.

\bibitem[Moazzam et~al.(2013)Moazzam, Kamal, Mathavan, Usman, and
  Rahman]{moazzam2013kinect}
Moazzam, I.; Kamal, K.; Mathavan, S.; Usman, S.; Rahman, M.
\newblock Metrology and visualization of potholes using the microsoft kinect
  sensor.
\newblock In Proceedings of the 16th International IEEE Conference on
  Intelligent Transportation Systems (ITSC 2013). IEEE,  2013, pp. 1284--1291.
\newblock {\url{https://doi.org/10.1109/itsc.2013.6728408}}.

\bibitem[Fan et~al.(2020)Fan, Ozgunalp, Hosking, Liu, and
  Pitas]{fan2020disparity}
Fan, R.; Ozgunalp, U.; Hosking, B.; Liu, M.; Pitas, I.
\newblock Pothole Detection Based on Disparity Transformation and Road Surface
  Modeling.
\newblock {\em IEEE Transactions on Image Processing} {\bf 2020}, {\em
  29},~897--908.
\newblock {\url{https://doi.org/10.1109/tip.2019.2933750}}.

\bibitem[Fan et~al.(2022)Fan, Ozgunalp, Wang, Liu, and
  Pitas]{fan2022rethinking}
Fan, R.; Ozgunalp, U.; Wang, Y.; Liu, M.; Pitas, I.
\newblock Rethinking Road Surface 3-D Reconstruction and Pothole Detection:
  From Perspective Transformation to Disparity Map Segmentation.
\newblock {\em IEEE Transactions on Cybernetics} {\bf 2022}, {\em
  52},~5799--5808.
\newblock {\url{https://doi.org/10.1109/tcyb.2021.3060461}}.

\bibitem[Dhiman and Klette(2020)]{dhiman2020pothole}
Dhiman, A.; Klette, R.
\newblock Pothole Detection Using Computer Vision and Learning.
\newblock {\em IEEE Transactions on Intelligent Transportation Systems} {\bf
  2020}, {\em 21},~3536--3550.
\newblock {\url{https://doi.org/10.1109/tits.2019.2931297}}.

\bibitem[Yurdakul and Tasdemir(2025)]{yurdakul2025yolov8}
Yurdakul, M.; Tasdemir, {\c{S}}.
\newblock An Enhanced YOLOv8 Model for Real-Time and Accurate Pothole Detection
  and Measurement.
\newblock {\em arXiv preprint arXiv:2505.04207} {\bf 2025}.

\bibitem[Abu-raddaha et~al.(2024)Abu-raddaha, El-Shair, and
  Rawashdeh]{aburaddaha2024jimaging}
Abu-raddaha, A.; El-Shair, Z.A.; Rawashdeh, S.
\newblock Leveraging Perspective Transformation for Enhanced Pothole Detection
  in Autonomous Vehicles.
\newblock {\em Journal of Imaging} {\bf 2024}, {\em 10},~227.
\newblock {\url{https://doi.org/10.3390/jimaging10090227}}.

\bibitem[Radford et~al.(2021)Radford, Kim, Hallacy, Ramesh, Goh, Agarwal,
  Sastry, Askell, et~al.]{radford2021clip}
Radford, A.; Kim, J.W.; Hallacy, C.; Ramesh, A.; Goh, G.; Agarwal, S.; Sastry,
  G.; Askell, A.;  et~al.
\newblock Learning Transferable Visual Models From Natural Language
  Supervision.
\newblock In Proceedings of the Proceedings of the 38th International
  Conference on Machine Learning (ICML),  2021.
\newblock arXiv:2103.00020.

\bibitem[Liu et~al.(2023)Liu, Li, Wu, and Lee]{liu2023llava}
Liu, H.; Li, C.; Wu, Q.; Lee, Y.J.
\newblock Visual Instruction Tuning.
\newblock In Proceedings of the Advances in Neural Information Processing
  Systems 36. Neural Information Processing Systems Foundation, Inc. (NeurIPS),
   2023, pp. 34892--34916.
\newblock {\url{https://doi.org/10.52202/075280-1516}}.

\bibitem[Comanici et~al.(2025)Comanici, Bieber, Schaekermann, Pasupat,
  Sachdeva, Dhillon, Blistein, Ram, et~al.]{comanici2025gemini25}
Comanici, G.; Bieber, E.; Schaekermann, M.; Pasupat, I.; Sachdeva, N.; Dhillon,
  I.; Blistein, M.; Ram, O.;  et~al.
\newblock Gemini 2.5: Pushing the Frontier with Advanced Reasoning,
  Multimodality, Long Context, and Next Generation Agentic Capabilities.
\newblock {\em arXiv preprint arXiv:2507.06261} {\bf 2025}.

\bibitem[Bai et~al.(2025{\natexlab{a}})Bai, Chen, Liu, Wang, Ge, Song, Dang,
  Wang, et~al.]{bai2025qwen25vl}
Bai, S.; Chen, K.; Liu, X.; Wang, J.; Ge, W.; Song, S.; Dang, K.; Wang, P.;
  et~al.
\newblock Qwen2.5-VL Technical Report.
\newblock {\em arXiv preprint arXiv:2502.13923} {\bf 2025}.

\bibitem[Bai et~al.(2025{\natexlab{b}})Bai, Cai, Chen, Chen, Chen, Cheng, Deng,
  Ding, et~al.]{bai2025qwen3vl}
Bai, S.; Cai, Y.; Chen, R.; Chen, K.; Chen, X.; Cheng, Z.; Deng, L.; Ding, W.;
  et~al.
\newblock Qwen3-VL Technical Report.
\newblock {\em arXiv preprint arXiv:2511.21631} {\bf 2025}.

\bibitem[Kim et~al.(2018)Kim, Rohrbach, Darrell, Canny, and
  Akata]{kim2018textual}
Kim, J.; Rohrbach, A.; Darrell, T.; Canny, J.; Akata, Z., Textual Explanations
  for Self-Driving Vehicles.
\newblock In {\em Computer Vision -- ECCV 2018}; Springer International
  Publishing,  2018; pp. 577--593.
\newblock {\url{https://doi.org/10.1007/978-3-030-01216-8_35}}.

\bibitem[Sima et~al.(2024)Sima, Renz, Chitta, Chen, Zhang, Xie, Bei{\ss}wenger,
  Luo, Geiger, and Li]{sima2024drivelm}
Sima, C.; Renz, K.; Chitta, K.; Chen, L.; Zhang, H.; Xie, C.; Bei{\ss}wenger,
  J.; Luo, P.; Geiger, A.; Li, H., DriveLM: Driving with Graph Visual Question
  Answering.
\newblock In {\em Computer Vision -- ECCV 2024}; Springer Nature Switzerland,
  2024; pp. 256--274.
\newblock {\url{https://doi.org/10.1007/978-3-031-72943-0_15}}.

\bibitem[Marcu et~al.(2024)Marcu, Chen, H{\"u}nermann, Karnsund, Hanotte,
  Chidananda, Nair, Badrinarayanan, Kendall, Shotton, Arani, and
  Sinavski]{marcu2024lingoqa}
Marcu, A.M.; Chen, L.; H{\"u}nermann, J.; Karnsund, A.; Hanotte, B.;
  Chidananda, P.; Nair, S.; Badrinarayanan, V.; Kendall, A.; Shotton, J.;
  et~al., LingoQA: Visual Question Answering for Autonomous Driving.
\newblock In {\em Computer Vision -- ECCV 2024}; Springer Nature Switzerland,
  2024; pp. 252--269.
\newblock {\url{https://doi.org/10.1007/978-3-031-72980-5_15}}.

\bibitem[Xu et~al.(2024)Xu, Zhang, Xie, Zhao, Guo, Wong, Li, and
  Zhao]{xu2024drivegpt4}
Xu, Z.; Zhang, Y.; Xie, E.; Zhao, Z.; Guo, Y.; Wong, K.Y.K.; Li, Z.; Zhao, H.
\newblock DriveGPT4: Interpretable End-to-End Autonomous Driving Via Large
  Language Model.
\newblock {\em IEEE Robotics and Automation Letters} {\bf 2024}, {\em
  9},~8186--8193.
\newblock {\url{https://doi.org/10.1109/lra.2024.3440097}}.

\bibitem[Tian et~al.(2024)Tian, Gu, Li, Liu, Wang, Zhao, Zhan, Jia,
  et~al.]{tian2024drivevlm}
Tian, X.; Gu, J.; Li, B.; Liu, Y.; Wang, Y.; Zhao, Z.; Zhan, K.; Jia, P.;
  et~al.
\newblock DriveVLM: The Convergence of Autonomous Driving and Large
  Vision-Language Models.
\newblock In Proceedings of the Proceedings of the Conference on Robot Learning
  (CoRL),  2024.
\newblock arXiv:2402.12289.

\bibitem[Cui et~al.(2024)Cui, Ma, Cao, Ye, Zhou, Liang, Chen, Lu, Yang, Liao,
  Gao, Li, Tang, Cao, Zhou, Liu, Yan, Mei, Cao, Wang, and Zheng]{cui2024survey}
Cui, C.; Ma, Y.; Cao, X.; Ye, W.; Zhou, Y.; Liang, K.; Chen, J.; Lu, J.; Yang,
  Z.; Liao, K.D.;  et~al.
\newblock A Survey on Multimodal Large Language Models for Autonomous Driving.
\newblock In Proceedings of the 2024 IEEE/CVF Winter Conference on Applications
  of Computer Vision Workshops (WACVW). IEEE,  2024, pp. 958--979.
\newblock {\url{https://doi.org/10.1109/wacvw60836.2024.00106}}.

\bibitem[Jongwiriyanurak et~al.(2025)Jongwiriyanurak, Zeng, Goo, Wang,
  Ilyankou, Sriroongvikrai, Christie, Wang, Chen, and
  Haworth]{jongwiriyanurak2024vroast}
Jongwiriyanurak, N.; Zeng, Z.; Goo, J.M.; Wang, X.; Ilyankou, I.;
  Sriroongvikrai, K.; Christie, N.; Wang, M.; Chen, H.; Haworth, J.
\newblock V-RoAst: Visual Road Assessment Can VLM be a Road Safety Assessor
  using the iRAP Standard?
\newblock In Proceedings of the 2025 IEEE/CVF International Conference on
  Computer Vision Workshops (ICCVW). IEEE,  2025, pp. 1669--1678.
\newblock {\url{https://doi.org/10.1109/iccvw69036.2025.00176}}.

\bibitem[Liu et~al.(2019)Liu, Liu, Tang, Yu, Wang, and Shi]{liu2019edge}
Liu, S.; Liu, L.; Tang, J.; Yu, B.; Wang, Y.; Shi, W.
\newblock Edge Computing for Autonomous Driving: Opportunities and Challenges.
\newblock {\em Proceedings of the IEEE} {\bf 2019}, {\em 107},~1697--1716.
\newblock {\url{https://doi.org/10.1109/jproc.2019.2915983}}.

\bibitem[Lin et~al.(2018)Lin, Zhang, Hsu, Skach, Haque, Tang, and
  Mars]{lin2018architectural}
Lin, S.C.; Zhang, Y.; Hsu, C.H.; Skach, M.; Haque, M.E.; Tang, L.; Mars, J.
\newblock The Architectural Implications of Autonomous Driving: Constraints and
  Acceleration.
\newblock In Proceedings of the Proceedings of the Twenty-Third International
  Conference on Architectural Support for Programming Languages and Operating
  Systems. ACM,  2018, pp. 751--766.
\newblock {\url{https://doi.org/10.1145/3173162.3173191}}.

\bibitem[Zhang et~al.(2023)Zhang, Han, Qiao, Kim, Bae, Lee, and
  Hong]{zhang2023mobilesam}
Zhang, C.; Han, D.; Qiao, Y.; Kim, J.U.; Bae, S.H.; Lee, S.; Hong, C.S.
\newblock Faster Segment Anything: Towards Lightweight SAM for Mobile
  Applications.
\newblock {\em arXiv preprint arXiv:2306.14289} {\bf 2023}.

\bibitem[Shu et~al.(2025)Shu, Li, Tang, Zhang, Chen, Li, Wang, and
  Chen]{shu2025tinysam}
Shu, H.; Li, W.; Tang, Y.; Zhang, Y.; Chen, Y.; Li, H.; Wang, Y.; Chen, X.
\newblock TinySAM: Pushing the Envelope for Efficient Segment Anything Model.
\newblock {\em Proceedings of the AAAI Conference on Artificial Intelligence}
  {\bf 2025}, {\em 39},~20470--20478.
\newblock {\url{https://doi.org/10.1609/aaai.v39i19.34255}}.

\bibitem[Zhang et~al.(2024)Zhang, Cai, and Han]{zhang2024efficientvitsam}
Zhang, Z.; Cai, H.; Han, S.
\newblock EfficientViT-SAM: Accelerated Segment Anything Model Without
  Performance Loss.
\newblock In Proceedings of the 2024 IEEE/CVF Conference on Computer Vision and
  Pattern Recognition Workshops (CVPRW). IEEE,  2024, pp. 7859--7863.
\newblock {\url{https://doi.org/10.1109/cvprw63382.2024.00782}}.

\bibitem[Zhou et~al.(2025)Zhou, Li, Loy, and Dai]{zhou2025edgesam}
Zhou, C.; Li, X.; Loy, C.C.; Dai, B.
\newblock EdgeSAM: Prompt-In-the-Loop Distillation for SAM.
\newblock {\em International Journal of Computer Vision} {\bf 2025}, {\em
  133},~8452--8468.
\newblock {\url{https://doi.org/10.1007/s11263-025-02562-9}}.

\bibitem[Xiong et~al.(2024)Xiong, Varadarajan, Wu, Xiang, Xiao, Zhu, Dai, Wang,
  Sun, Iandola, Krishnamoorthi, and Chandra]{xiong2024efficientsam}
Xiong, Y.; Varadarajan, B.; Wu, L.; Xiang, X.; Xiao, F.; Zhu, C.; Dai, X.;
  Wang, D.; Sun, F.; Iandola, F.;  et~al.
\newblock EfficientSAM: Leveraged Masked Image Pretraining for Efficient
  Segment Anything.
\newblock In Proceedings of the 2024 IEEE/CVF Conference on Computer Vision and
  Pattern Recognition (CVPR). IEEE,  2024, pp. 16111--16121.
\newblock {\url{https://doi.org/10.1109/cvpr52733.2024.01525}}.

\bibitem[Zhao et~al.(2023)Zhao, Ding, An, Du, Yu, Li, Tang, and
  Wang]{zhao2023fastsam}
Zhao, X.; Ding, W.; An, Y.; Du, Y.; Yu, T.; Li, M.; Tang, M.; Wang, J.
\newblock Fast Segment Anything.
\newblock {\em arXiv preprint arXiv:2306.12156} {\bf 2023}.

\bibitem[Ravi et~al.(2025)Ravi, Gabeur, Hu, Hu, Ryali, Ma, Khedr, R{\"a}dle,
  et~al.]{ravi2024sam2}
Ravi, N.; Gabeur, V.; Hu, Y.T.; Hu, R.; Ryali, C.; Ma, T.; Khedr, H.;
  R{\"a}dle, R.;  et~al.
\newblock SAM 2: Segment Anything in Images and Videos.
\newblock In Proceedings of the Proceedings of the International Conference on
  Learning Representations (ICLR),  2025.
\newblock arXiv:2408.00714.

\bibitem[Mazurowski et~al.(2023)Mazurowski, Dong, Gu, Yang, Konz, and
  Zhang]{mazurowski2023medsam}
Mazurowski, M.A.; Dong, H.; Gu, H.; Yang, J.; Konz, N.; Zhang, Y.
\newblock Segment anything model for medical image analysis: An experimental
  study.
\newblock {\em Medical Image Analysis} {\bf 2023}, {\em 89},~102918.
\newblock {\url{https://doi.org/10.1016/j.media.2023.102918}}.

\bibitem[Ji et~al.(2024)Ji, Li, Bi, Liu, Li, and Cheng]{ji2024sam_not_perfect}
Ji, W.; Li, J.; Bi, Q.; Liu, T.; Li, W.; Cheng, L.
\newblock Segment Anything Is Not Always Perfect: An Investigation of SAM on
  Different Real-world Applications.
\newblock {\em Machine Intelligence Research} {\bf 2024}, {\em 21},~617--630.
\newblock {\url{https://doi.org/10.1007/s11633-023-1385-0}}.

\bibitem[Micikevicius et~al.(2018)Micikevicius, Narang, Alben, Diamos, Elsen,
  Garcia, Ginsburg, Houston, et~al.]{micikevicius2018mixed}
Micikevicius, P.; Narang, S.; Alben, J.; Diamos, G.; Elsen, E.; Garcia, D.;
  Ginsburg, B.; Houston, M.;  et~al.
\newblock Mixed Precision Training.
\newblock In Proceedings of the Proceedings of the International Conference on
  Learning Representations (ICLR),  2018.
\newblock arXiv:1710.03740.

\bibitem[Ansel et~al.(2024)Ansel, Yang, He, Gimelshein, Jain, Voznesensky, Bao,
  Bell, Berard, Burovski, Chauhan, Chourdia, Constable, Desmaison, DeVito,
  Ellison, Feng, Gong, Gschwind, Hirsh, Huang, Kalambarkar, Kirsch, Lazos,
  Lezcano, Liang, Liang, Lu, Luk, Maher, Pan, Puhrsch, Reso, Saroufim,
  Siraichi, Suk, Zhang, Suo, Tillet, Zhao, Wang, Zhou, Zou, Wang, Mathews, Wen,
  Chanan, Wu, and Chintala]{ansel2024pytorch2}
Ansel, J.; Yang, E.; He, H.; Gimelshein, N.; Jain, A.; Voznesensky, M.; Bao,
  B.; Bell, P.; Berard, D.; Burovski, E.;  et~al.
\newblock PyTorch 2: Faster Machine Learning Through Dynamic Python Bytecode
  Transformation and Graph Compilation.
\newblock In Proceedings of the Proceedings of the 29th ACM International
  Conference on Architectural Support for Programming Languages and Operating
  Systems, Volume 2. ACM,  2024, pp. 929--947.
\newblock {\url{https://doi.org/10.1145/3620665.3640366}}.

\bibitem[Hinton et~al.(2015)Hinton, Vinyals, and Dean]{hinton2015distilling}
Hinton, G.; Vinyals, O.; Dean, J.
\newblock Distilling the Knowledge in a Neural Network.
\newblock {\em arXiv preprint arXiv:1503.02531} {\bf 2015}.

\bibitem[Gou et~al.(2021)Gou, Yu, Maybank, and Tao]{gou2021kdsurvey}
Gou, J.; Yu, B.; Maybank, S.J.; Tao, D.
\newblock Knowledge Distillation: A Survey.
\newblock {\em International Journal of Computer Vision} {\bf 2021}, {\em
  129},~1789--1819.
\newblock {\url{https://doi.org/10.1007/s11263-021-01453-z}}.

\bibitem[Xie et~al.(2020)Xie, Luong, Hovy, and Le]{xie2020noisystudent}
Xie, Q.; Luong, M.T.; Hovy, E.; Le, Q.V.
\newblock Self-Training With Noisy Student Improves ImageNet Classification.
\newblock In Proceedings of the 2020 IEEE/CVF Conference on Computer Vision and
  Pattern Recognition (CVPR). IEEE,  2020, pp. 10684--10695.
\newblock {\url{https://doi.org/10.1109/cvpr42600.2020.01070}}.

\bibitem[Hsieh et~al.(2023)Hsieh, Li, Yeh, Nakhost, Fujii, Ratner, Krishna,
  Lee, and Pfister]{hsieh2023stepbystep}
Hsieh, C.Y.; Li, C.L.; Yeh, C.k.; Nakhost, H.; Fujii, Y.; Ratner, A.; Krishna,
  R.; Lee, C.Y.; Pfister, T.
\newblock Distilling Step-by-Step! Outperforming Larger Language Models with
  Less Training Data and Smaller Model Sizes.
\newblock In Proceedings of the Findings of the Association for Computational
  Linguistics: ACL 2023. Association for Computational Linguistics,  2023, pp.
  8003--8017.
\newblock {\url{https://doi.org/10.18653/v1/2023.findings-acl.507}}.

\bibitem[Geirhos et~al.(2020)Geirhos, Jacobsen, Michaelis, Zemel, Brendel,
  Bethge, and Wichmann]{geirhos2020shortcut}
Geirhos, R.; Jacobsen, J.H.; Michaelis, C.; Zemel, R.; Brendel, W.; Bethge, M.;
  Wichmann, F.A.
\newblock Shortcut learning in deep neural networks.
\newblock {\em Nature Machine Intelligence} {\bf 2020}, {\em 2},~665--673.
\newblock {\url{https://doi.org/10.1038/s42256-020-00257-z}}.

\bibitem[Recht et~al.(2019)Recht, Roelofs, Schmidt, and
  Shankar]{recht2019imagenet}
Recht, B.; Roelofs, R.; Schmidt, L.; Shankar, V.
\newblock Do ImageNet Classifiers Generalize to ImageNet?
\newblock In Proceedings of the Proceedings of the 36th International
  Conference on Machine Learning (ICML),  2019.
\newblock arXiv:1902.10811.

\bibitem[Koh et~al.(2021)Koh, Sagawa, Marklund, Xie, Zhang, Balsubramani, Hu,
  Yasunaga, et~al.]{koh2021wilds}
Koh, P.W.; Sagawa, S.; Marklund, H.; Xie, S.M.; Zhang, M.; Balsubramani, A.;
  Hu, W.; Yasunaga, M.;  et~al.
\newblock WILDS: A Benchmark of in-the-Wild Distribution Shifts.
\newblock In Proceedings of the Proceedings of the 38th International
  Conference on Machine Learning (ICML),  2021.
\newblock arXiv:2012.07421.

\bibitem[Lakshminarayanan et~al.(2017)Lakshminarayanan, Pritzel, and
  Blundell]{lakshminarayanan2017ensembles}
Lakshminarayanan, B.; Pritzel, A.; Blundell, C.
\newblock Simple and Scalable Predictive Uncertainty Estimation using Deep
  Ensembles.
\newblock In Proceedings of the Advances in Neural Information Processing
  Systems 30 (NeurIPS),  2017.
\newblock arXiv:1612.01474.

\bibitem[Liu et~al.(2022)Liu, Mao, Wu, Feichtenhofer, Darrell, and
  Xie]{liu2022convnext}
Liu, Z.; Mao, H.; Wu, C.Y.; Feichtenhofer, C.; Darrell, T.; Xie, S.
\newblock A ConvNet for the 2020s.
\newblock In Proceedings of the 2022 IEEE/CVF Conference on Computer Vision and
  Pattern Recognition (CVPR). IEEE,  2022, pp. 11966--11976.
\newblock {\url{https://doi.org/10.1109/cvpr52688.2022.01167}}.

\bibitem[Willard and Louf(2023)]{willard2023outlines}
Willard, B.T.; Louf, R.
\newblock Efficient Guided Generation for Large Language Models.
\newblock {\em arXiv preprint arXiv:2307.09702} {\bf 2023}.

\bibitem[Geng et~al.(2023)Geng, Josifoski, Peyrard, and West]{geng2023grammar}
Geng, S.; Josifoski, M.; Peyrard, M.; West, R.
\newblock Grammar-Constrained Decoding for Structured NLP Tasks without
  Finetuning.
\newblock In Proceedings of the Proceedings of the 2023 Conference on Empirical
  Methods in Natural Language Processing. Association for Computational
  Linguistics,  2023, pp. 10932--10952.
\newblock {\url{https://doi.org/10.18653/v1/2023.emnlp-main.674}}.

\bibitem[Dong et~al.(2024)Dong, Ruan, Cai, Lai, Xu, Zhao, and
  Chen]{dong2024xgrammar}
Dong, Y.; Ruan, C.F.; Cai, Y.; Lai, R.; Xu, Z.; Zhao, Y.; Chen, T.
\newblock XGrammar: Flexible and Efficient Structured Generation Engine for
  Large Language Models.
\newblock {\em arXiv preprint arXiv:2411.15100} {\bf 2024}.

\bibitem[Kwon et~al.(2023)Kwon, Li, Zhuang, Sheng, Zheng, Yu, Gonzalez, Zhang,
  and Stoica]{kwon2023vllm}
Kwon, W.; Li, Z.; Zhuang, S.; Sheng, Y.; Zheng, L.; Yu, C.H.; Gonzalez, J.;
  Zhang, H.; Stoica, I.
\newblock Efficient Memory Management for Large Language Model Serving with
  PagedAttention.
\newblock In Proceedings of the Proceedings of the 29th Symposium on Operating
  Systems Principles. ACM,  2023, pp. 611--626.
\newblock {\url{https://doi.org/10.1145/3600006.3613165}}.

\bibitem[Zheng et~al.(2024)Zheng, Yin, Xie, Sun, Huang, Yu, Cao, Kozyrakis,
  Stoica, Gonzalez, Barrett, and Sheng]{zheng2024sglang}
Zheng, L.; Yin, L.; Xie, Z.; Sun, C.; Huang, J.; Yu, C.; Cao, S.; Kozyrakis,
  C.; Stoica, I.; Gonzalez, J.;  et~al.
\newblock SGLang: Efficient Execution of Structured Language Model Programs.
\newblock In Proceedings of the Advances in Neural Information Processing
  Systems 37. Neural Information Processing Systems Foundation, Inc. (NeurIPS),
   2024, pp. 62557--62583.
\newblock {\url{https://doi.org/10.52202/079017-2000}}.

\bibitem[Tam et~al.(2024)Tam, Wu, Tsai, Lin, Lee, and Chen]{tam2024speakfreely}
Tam, Z.R.; Wu, C.K.; Tsai, Y.L.; Lin, C.Y.; Lee, H.y.; Chen, Y.N.
\newblock Let Me Speak Freely? A Study On The Impact Of Format Restrictions On
  Large Language Model Performance.
\newblock In Proceedings of the Proceedings of the 2024 Conference on Empirical
  Methods in Natural Language Processing: Industry Track. Association for
  Computational Linguistics,  2024, pp. 1218--1236.
\newblock {\url{https://doi.org/10.18653/v1/2024.emnlp-industry.91}}.

\bibitem[Wei et~al.(2022)Wei, Wang, Schuurmans, Bosma, Ichter, Xia, Chi, Le,
  and Zhou]{wei2022cot}
Wei, J.; Wang, X.; Schuurmans, D.; Bosma, M.; Ichter, B.; Xia, F.; Chi, E.; Le,
  Q.V.; Zhou, D.
\newblock Chain-Of-Thought Prompting Elicits Reasoning in Large Language
  Models.
\newblock In Proceedings of the Advances in Neural Information Processing
  Systems 35. Neural Information Processing Systems Foundation, Inc. (NeurIPS),
   2022, pp. 24824--24837.
\newblock {\url{https://doi.org/10.52202/068431-1800}}.

\bibitem[Cohen(1960)]{cohen1960}
Cohen, J.
\newblock A Coefficient of Agreement for Nominal Scales.
\newblock {\em Educational and Psychological Measurement} {\bf 1960}, {\em
  20},~37--46.
\newblock {\url{https://doi.org/10.1177/001316446002000104}}.

\bibitem[Fleiss(1971)]{fleiss1971}
Fleiss, J.L.
\newblock Measuring nominal scale agreement among many raters.
\newblock {\em Psychological Bulletin} {\bf 1971}, {\em 76},~378--382.
\newblock {\url{https://doi.org/10.1037/h0031619}}.

\bibitem[Shrout and Fleiss(1979)]{shrout1979icc}
Shrout, P.E.; Fleiss, J.L.
\newblock Intraclass correlations: Uses in assessing rater reliability.
\newblock {\em Psychological Bulletin} {\bf 1979}, {\em 86},~420--428.
\newblock {\url{https://doi.org/10.1037/0033-2909.86.2.420}}.

\bibitem[Koo and Li(2016)]{koo2016icc}
Koo, T.K.; Li, M.Y.
\newblock A Guideline of Selecting and Reporting Intraclass Correlation
  Coefficients for Reliability Research.
\newblock {\em Journal of Chiropractic Medicine} {\bf 2016}, {\em
  15},~155--163.
\newblock {\url{https://doi.org/10.1016/j.jcm.2016.02.012}}.

\bibitem[Hayes and Krippendorff(2007)]{hayes2007kripp}
Hayes, A.F.; Krippendorff, K.
\newblock Answering the Call for a Standard Reliability Measure for Coding
  Data.
\newblock {\em Communication Methods and Measures} {\bf 2007}, {\em 1},~77--89.
\newblock {\url{https://doi.org/10.1080/19312450709336664}}.

\bibitem[Zheng et~al.(2023)Zheng, Chiang, Sheng, Zhuang, Wu, Zhuang, Lin, Li,
  Li, Xing, Zhang, Gonzalez, and Stoica]{zheng2023judging}
Zheng, L.; Chiang, W.L.; Sheng, Y.; Zhuang, S.; Wu, Z.; Zhuang, Y.; Lin, Z.;
  Li, Z.; Li, D.; Xing, E.;  et~al.
\newblock Judging LLM-as-a-Judge with MT-Bench and Chatbot Arena.
\newblock In Proceedings of the Advances in Neural Information Processing
  Systems 36. Neural Information Processing Systems Foundation, Inc. (NeurIPS),
   2023, pp. 46595--46623.
\newblock {\url{https://doi.org/10.52202/075280-2020}}.

\bibitem[Chiang and Lee(2023)]{chiang2023human}
Chiang, C.H.; Lee, H.y.
\newblock Can Large Language Models Be an Alternative to Human Evaluations?
\newblock In Proceedings of the Proceedings of the 61st Annual Meeting of the
  Association for Computational Linguistics (Volume 1: Long Papers).
  Association for Computational Linguistics,  2023, pp. 15607--15631.
\newblock {\url{https://doi.org/10.18653/v1/2023.acl-long.870}}.

\bibitem[Chen et~al.(2024)Chen, Chen, Zhang, Liu, Wang, Zhou, Zhang, Wan,
  et~al.]{chen2024mllmjudge}
Chen, D.; Chen, R.; Zhang, S.; Liu, Y.; Wang, Y.; Zhou, H.; Zhang, Q.; Wan, Y.;
   et~al.
\newblock MLLM-as-a-Judge: Assessing Multimodal LLM-as-a-Judge with
  Vision-Language Benchmark.
\newblock In Proceedings of the Proceedings of the 41st International
  Conference on Machine Learning (ICML),  2024.
\newblock arXiv:2402.04788.

\bibitem[Wang et~al.(2024)Wang, Li, Chen, Cai, Zhu, Lin, Cao, Kong, Liu, Liu,
  and Sui]{wang2024notfair}
Wang, P.; Li, L.; Chen, L.; Cai, Z.; Zhu, D.; Lin, B.; Cao, Y.; Kong, L.; Liu,
  Q.; Liu, T.;  et~al.
\newblock Large Language Models are not Fair Evaluators.
\newblock In Proceedings of the Proceedings of the 62nd Annual Meeting of the
  Association for Computational Linguistics (Volume 1: Long Papers).
  Association for Computational Linguistics,  2024, pp. 9440--9450.
\newblock {\url{https://doi.org/10.18653/v1/2024.acl-long.511}}.

\bibitem[Panickssery et~al.(2024)Panickssery, Bowman, and
  Feng]{panickssery2024selfpref}
Panickssery, A.; Bowman, S.; Feng, S.
\newblock LLM Evaluators Recognize and Favor Their Own Generations.
\newblock In Proceedings of the Advances in Neural Information Processing
  Systems 37. Neural Information Processing Systems Foundation, Inc. (NeurIPS),
   2024, pp. 68772--68802.
\newblock {\url{https://doi.org/10.52202/079017-2197}}.

\bibitem[Bavaresco et~al.(2025)Bavaresco, Bernardi, Bertolazzi, Elliott,
  Fern{\'a}ndez, Gatt, Ghaleb, Giulianelli, Hanna, Koller, Martins, Mondorf,
  Neplenbroek, Pezzelle, Plank, Schlangen, Suglia, Surikuchi, Takmaz, and
  Testoni]{bavaresco2025judges}
Bavaresco, A.; Bernardi, R.; Bertolazzi, L.; Elliott, D.; Fern{\'a}ndez, R.;
  Gatt, A.; Ghaleb, E.; Giulianelli, M.; Hanna, M.; Koller, A.;  et~al.
\newblock LLMs instead of Human Judges? A Large Scale Empirical Study across 20
  NLP Evaluation Tasks.
\newblock In Proceedings of the Proceedings of the 63rd Annual Meeting of the
  Association for Computational Linguistics (Volume 2: Short Papers).
  Association for Computational Linguistics,  2025, pp. 238--255.
\newblock {\url{https://doi.org/10.18653/v1/2025.acl-short.20}}.

\bibitem[Yurdakul and Ta{\c{s}}demir(2025)]{pothrgbd_data}
Yurdakul, M.; Ta{\c{s}}demir, {\c{S}}.
\newblock {PothRGBD}: {RGB} and Depth Images of Potholes.
\newblock IEEE DataPort,  2025.
\newblock {\url{https://doi.org/10.21227/z8eq-sf60}}.

\bibitem[McNemar(1947)]{mcnemar1947}
McNemar, Q.
\newblock Note on the Sampling Error of the Difference Between Correlated
  Proportions or Percentages.
\newblock {\em Psychometrika} {\bf 1947}, {\em 12},~153--157.
\newblock {\url{https://doi.org/10.1007/bf02295996}}.

\bibitem[Wilcoxon(1945)]{wilcoxon1945}
Wilcoxon, F.
\newblock Individual Comparisons by Ranking Methods.
\newblock {\em Biometrics Bulletin} {\bf 1945}, {\em 1},~80.
\newblock {\url{https://doi.org/10.2307/3001968}}.

\bibitem[Efron(1979)]{efron1979bootstrap}
Efron, B.
\newblock Bootstrap Methods: Another Look at the Jackknife.
\newblock {\em The Annals of Statistics} {\bf 1979}, {\em 7},~1--26.
\newblock {\url{https://doi.org/10.1214/aos/1176344552}}.

\end{thebibliography}
\end{adjustwidth}

\PublishersNote{}
\end{document}